\documentclass[letterpaper]{article} 
\usepackage{aaai2026}  
\usepackage{times}  
\usepackage{helvet}  
\usepackage{courier}  
\usepackage[hyphens]{url}  
\usepackage{graphicx} 
\usepackage{natbib}  
\usepackage{caption} 
\usepackage{algorithm}
\usepackage{algorithmic}
\usepackage{fvextra}

\usepackage{subcaption}

\usepackage{hyperref}
\usepackage{soul}
\usepackage{booktabs}
\usepackage{colortbl}
\usepackage{inconsolata}      
\usepackage{tcolorbox}         
\usepackage{enumitem}          
\usepackage{xcolor}         
\usepackage{authblk}
\usepackage{amsmath}
\usepackage{amssymb}
\usepackage{cleveref}
\usepackage{multirow}
\usepackage{caption}
\usepackage{graphicx}
\definecolor{codeblue}{RGB}{0, 82, 147}     
\definecolor{codegreen}{RGB}{0, 128, 0}    
\definecolor{codegray}{RGB}{100, 100, 100}  
\definecolor{codeorange}{RGB}{230, 145, 56} 
\definecolor{darkerblue}{rgb}{0,0.08,0.45} 
\definecolor{royalblue}{RGB}{65,105,225}
\definecolor{lightblue}{RGB}{221,235,247}
\definecolor{fig3blue}{RGB}{47, 122, 232}  
\definecolor{fig3red}{RGB}{213, 32, 52}
\definecolor{fig3green}{RGB}{0, 137, 72} 
\definecolor{fig3yellow}{RGB}{217, 161, 5}
\definecolor{gray94}{gray}{.94}
\definecolor{gray90}{gray}{.90}

\definecolor{darkgreen}{RGB}{34,139,34}
\newcommand{\green}[1]{\textcolor{darkgreen}{#1}}
\newcommand{\gray}[1]{\textcolor{gray}{#1}}
\newcommand{\gbf}[1]{\green{\bf{#1}}}

\definecolor{citecolor}{HTML}{2980b9}
\definecolor{linkcolor}{HTML}{c0392b}
\definecolor{lightblue}{RGB}{115, 192, 222}
\definecolor{lightgreen}{RGB}{145, 204, 117}
\newtcolorbox{promptbox}[3][Judge Prompt]{
colback=black!5!white,
arc=5pt, 
boxrule=0.5pt,
fonttitle=\bfseries,
title=#1, 
before upper={\small}, fontupper=\fontfamily{ptm}\selectfont,
colframe=#2,
label=#3,
}
\definecolor{lightorange}{RGB}{253, 208, 162} 

\newtcolorbox{promptbox*}[3][Judge Prompt]{
  float*=ht,              
  width=\textwidth,       
  colback=black!5,
  colframe=#2,
  arc=5pt,
  boxrule=0.5pt,
  fonttitle=\bfseries,
  title=#1,
  before upper={\small},
  fontupper=\fontfamily{ptm}\selectfont,
  label=#3,
}

\usepackage{newfloat}
\usepackage{listings}
\DeclareCaptionStyle{ruled}{labelfont=normalfont,labelsep=colon,strut=off} 
\floatstyle{ruled}
\newfloat{listing}{tb}{lst}{}
\floatname{listing}{Listing}

\title{Look-Back: Implicit Visual Re-focusing in MLLM Reasoning}
\author{%
  \textbf{Shuo Yang}\textsuperscript{1,*},
  \textbf{Yuwei Niu}\textsuperscript{1,*},
  \textbf{Yuyang Liu}\textsuperscript{1,$^\dagger$},
  \textbf{Yang Ye}\textsuperscript{1},
  \textbf{Bin Lin}\textsuperscript{1},
  \textbf{Li Yuan}\textsuperscript{1,2,$^\dagger$}
}

\affil{
  {\tt 
  $*$ Equal Contributors, $\dagger$ Corresponding Authors
  }
  \par
  \textsuperscript{1} Peking University, Shenzhen Graduate School, 
  \textsuperscript{2} Peng Cheng Laboratory
  \\
  {\tt 
  \{shuo$\_$yang@stu, yuanli-ece@\}.pku.edu.cn
  }
}

\definecolor{cornellred}{rgb}{0.7, 0.11, 0.11}

\definecolor{cadmiumgreen}{rgb}{0.0, 0.42, 0.24}

\definecolor{aliceblue}{rgb}{0.91, 0.94, 0.97}

\definecolor{darkblue}{rgb}{0.83, 0.89, 0.97}

\definecolor{Red7}{rgb}{0.941, 0.243, 0.243}

\definecolor{Green7}{RGB}{55, 178, 77}

\definecolor{Blue9}{rgb}{0.098,0.3,0.9}

\usepackage{pifont}

\sethlcolor{aliceblue}

\hypersetup{
 linkcolor = cornellred,
 citecolor = cadmiumgreen,
 colorlinks = true,
 urlcolor = Blue9
}

\begin{document}

\maketitle

\begin{abstract}
Multimodal Large Language Models (MLLMs) have achieved remarkable progress in multimodal reasoning. However, they often excessively rely on textual information during the later stages of inference, neglecting the crucial integration of visual input. Current methods typically address this by explicitly injecting visual information to guide the reasoning process. In this work, through an analysis of MLLM attention patterns, we made an intriguing observation: with appropriate guidance, MLLMs can spontaneously re-focus their attention on visual inputs during the later stages of reasoning, even without explicit visual information injection. This spontaneous shift in focus suggests that MLLMs are intrinsically capable of performing visual fusion reasoning. Building on this insight, we introduce Look-Back, an implicit approach designed to guide MLLMs to ``look back" at visual information in a self-directed manner during reasoning. Look-Back empowers the model to autonomously determine when, where, and how to re-focus on visual inputs, eliminating the need for explicit model-structure constraints or additional input. We demonstrate that Look-Back significantly enhances the model's reasoning and perception capabilities, as evidenced by extensive empirical evaluations on multiple multimodal benchmarks. \footnote{Code and models will be released at \url{https://github.com/PKU-YuanGroup/Look-Back}.}

\end{abstract}

\section{Introduction}
\label{sec:introduction}

With the development of multimodal reasoning \cite{amizadeh2020neuro, garcez2019neural, gupta2023visual, thawakar2025llamav, guo2024mammoth, bai2023qwen,hurst2024gpt, xu2024llava} and reinforcement learning with verifiable rewards (RLVR)~\cite{shao2024deepseekmath, guo2025deepseek, meng2025mmeureka, peng2025lmm}, Multimodal Large Language Models (MLLMs)~\cite{liu2023visual,qwen2.5-VL,Qwen2VL,liao2025langbridge,lin2025uniworld,wan2025meda} have made significant progress in jointly processing image and text inputs to perform complex tasks~\cite{google2025gemini,openai2025o3o4systemcard, jaech2024openai,pang2024next}. However, recent research indicates that most approaches still predominantly rely on text during the later stages of reasoning, neglecting the visual modality~\cite{zheng2025deepeyes, fan2025grit, su2025openthinkimg, zhang2025chain, yang2025r1, hu2024visual, liu2025visual, zou2024look}. Specifically, during the reasoning process, the model's attention to visual information gradually diminishes, almost reaching zero in the later stages~\cite{sun2025mitigating, tu2025attention, chen2024image}, to the extent that visual information in the later phases exerts negligible influence on the reasoning result~\cite {sun2025mitigating}. 

However, humans naturally integrate visual and cognitive processing in multimodal reasoning~\cite{najemnik2005optimal,tversky2002animation, tversky2005functional, kosslyn1996image, goel1995sketches, larkin1987diagram, zhang1994representations}, and OpenAI's o3~\cite{openai2025o3o4systemcard} represents the gradual shift in the field from solely text-based reasoning to deep integration with visual information. Despite this progress, most existing methods still \textbf{explicitly inject visual information}~\cite{zheng2025deepeyes, su2025openthinkimg, zhang2025chain, wang2025visuothink, chern2025thinking}, such as re-inputting images or re-injecting image tokens into the model~\cite{sarch2025grounded, wu2025reinforcing, xu2025visual, zhang2025mllms, gupta2023visual}. These methods essentially guide the model to re-focus its attention on visual cues. Based on this, we propose a critical research question:

\begin{center}
    \textit{Instead of explicitly re-injecting visual information, can MLLMs be enabled to self-directively and implicitly learn when and how to re-focus on visual input?}
\end{center}

Based on the aforementioned question, we conducted a preliminary experiment to validate that the model can autonomously re-focus on the image. Specifically, we introduced a simple prompt (as shown in Figure~\ref{fig:2}) into the original CoT framework. Surprisingly, the model spontaneously enhanced its attention to the image during the later stages of reasoning, re-focusing on the visual input without any additional explicit inputs or model-structure constraints.

\begin{figure*}[!t] 
    \centering
    \includegraphics[width=0.95\textwidth]{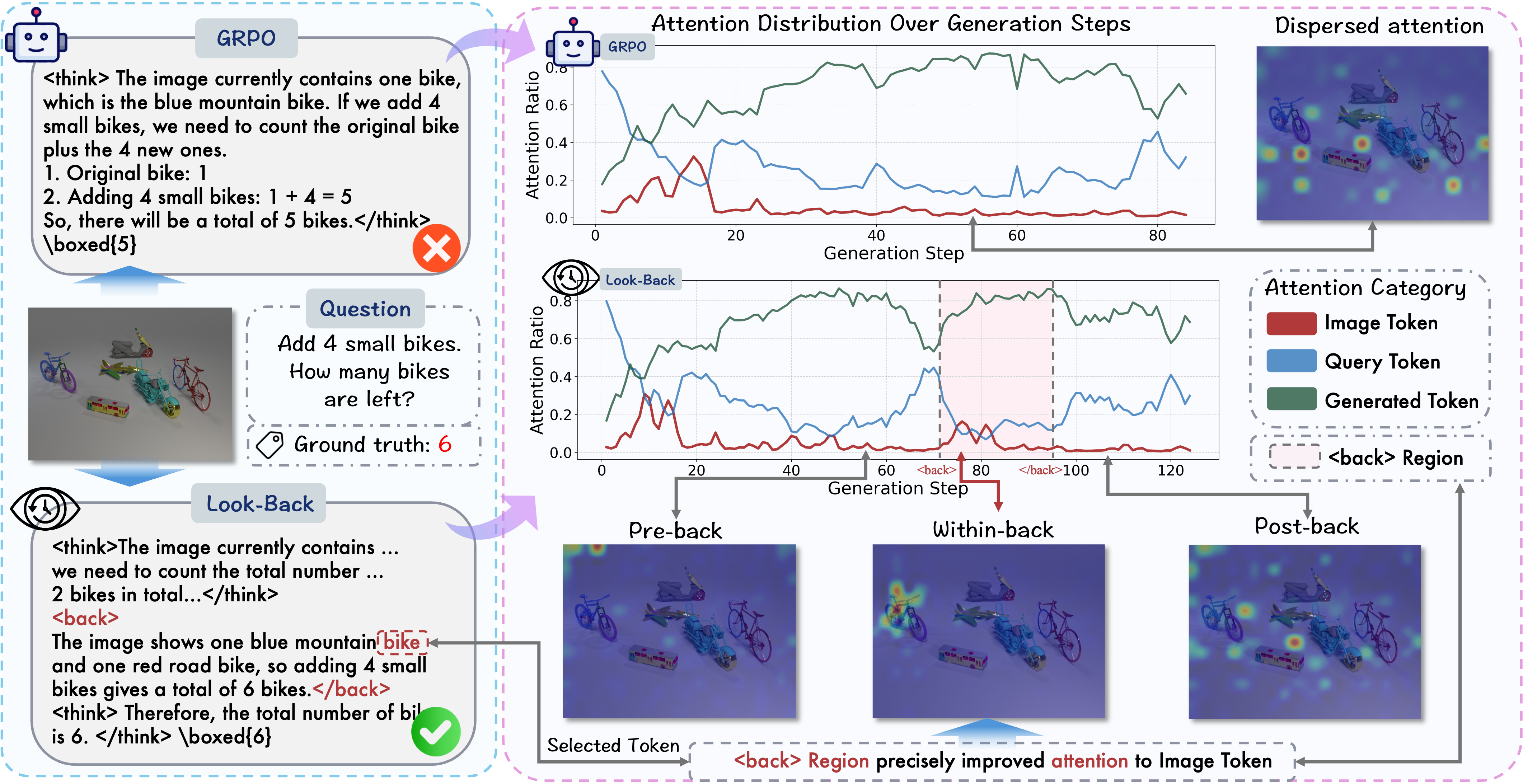} 
    \caption{An Overview of the Look-Back Mechanism. This figure contrasts the standard GRPO model with our Look-Back approach. The GRPO model (top left) miscounts the bikes due to diminished visual attention in later reasoning stages. In contrast, the Look-Back model (bottom left) utilizes a \texttt{<back>} token to re-focus on the image, correcting the initial count and reaching the right answer. The attention graphs show that Look-Back significantly increases attention to image tokens (red line) during the \texttt{<back>} phase, a behavior absent in GRPO. The attention maps below confirm that this re-focused attention is precisely targeted at the relevant objects in the image.}
    \label{fig:intro}
\end{figure*}

To better leverage the phenomenon of the model’s spontaneous attention to the image, we propose the \textbf{Look-Back} method, designed to guide MLLMs to ``look back" at the visual information during the reasoning process in a natural and self-directed manner, thus enhancing their attention to visual input. Specifically, we developed a two-stage training framework. In the first stage, we utilize advanced MLLMs to generate reflective data with the \texttt{<back>} token, followed by cold-start fine-tuning to lay the foundation for subsequent reinforcement learning training. In the second stage, We only introduce a format reward based on the \texttt{<back>} token for the GRPO algorithm, with the aim of further reinforcing the model’s ability to focus on visual information through reinforcement learning.

As shown in Figure\ref{fig:intro}, Look-Back effectively encourages MLLMs to spontaneously generate reflective reasoning content related to the image without explicitly injecting visual information, autonomously enhances attention to the image during the later stages of reasoning (i.e., re-focusing on the image). Through analysis of the attention maps, we confirmed that the model indeed attended to the correct visual location within the \texttt{<back>} token. Look-Back enables the model to autonomously decide \textbf{when} (the timing of triggering the \texttt{<back>} token is determined by the model), \textbf{where} (selecting specific regions of the image to attend to), and \textbf{how} (autonomously determining how to enhance attention) to reflect on visual input, all without requiring explicit inputs or structural constraints on the model.

This paper aims to propose an implicit visual fusion reasoning paradigm, generated spontaneously by the model, rather than merely evaluating which paradigm is the most effective. We conducted comprehensive experimental validation using the Qwen-2.5-VL-7B model~\cite{qwen2.5-VL} on multiple widely used multimodal reasoning benchmarks. The results indicate that by guiding the model to spontaneously re-focus on the image Look-Back can consistently enhance performance in reasoning and perception tasks. Our key contributions are summarized as follows:

\begin{itemize}
    \item By analyzing the trend of attention changes, we found that, without explicitly injecting visual information, the existing MLLM can autonomously attend to visual input.
    \item We introduced the Look-Back implicit training paradigm, which, after cold-start fine-tuning, can trigger the model’s visual reflection behavior by simply modifying the format reward function.
    \item Extensive evaluation on multiple multimodal benchmarks demonstrated that, Look-Back can consistently enhance performance in reasoning and perception tasks.
\end{itemize}

\section{Do MLLMs Know When and How to Reflect on Visual Input?}
Recent research~\cite{hu2024visual, zhang2025chain, su2025openthinkimg, fan2025grit,liu2025visual, zheng2025deepeyes} has revealed that Multimodal Large Language Models (MLLMs) often excessively rely on textual information during later stages of inference, neglecting the crucial integration of visual input. This diminishing attention to visual information as reasoning progresses significantly impacts the reliability and performance of vision-language models. Current approaches typically address this by explicitly injecting visual information to guide the reasoning process, such as re-inputting images into the model.

\begin{table*}[t]
 \centering
 \caption{Performance of Qwen-2.5-VL-7B on Math-Benchmark using different prompts. "CoT prompt" refers to the standard Chain-of-Thought prompt. "Back prompt" indicates our prompt that encourages the model to re-focus on the image within \texttt{<back>} tokens. "Trigger rate" denotes the percentage of inference responses where the model spontaneously generated the \texttt{<back>} token.}
 \label{table:7b_results_initial_experiment} 
 \begin{tabular}{lcccccc}
 \toprule[1.2pt]
 \textbf{Model} & MathVerse & MathVision & MathVista & WeMath & GeoMath & Avg\textsubscript{M} \\
 \midrule
 CoT prompt & 45.71 & 25.49 & 64.2 & 60.46 & 45.61 & 48.294 \\
 Back prompt & 46.62 & 26.58 & 67.3 & 63.22 & 47.93 & 50.33 \\
 \midrule
 Trigger rate & 62.01\% & 56.05\% & 87.00\% & 51.26\% & 56.10\% & 62.48\% \\
 \bottomrule[1.2pt]
 \end{tabular}
\end{table*}

\begin{table*}[t]
   \centering
\caption{Performance on Math-Benchmark comparing models \textit{with} and \textit{without} the \texttt{<back>} mechanism, specifically for instances where the \texttt{<back>} token was triggered. "w/o \textit{back}" represents the baseline performance, while "w/ \textit{back}" shows the performance when the model engages in visual reflection. $\Delta$ \textit{Gain} shows the percentage performance increase.}
\label{table:7b_results_with_back_gain} 
   \begin{tabular}{lcccccc}
     \toprule[1.2pt]
     \textbf{Model} & MathVerse & MathVision & MathVista & WeMath & GeoMath & Avg\textsubscript{M} \\
     \midrule

     w/o \textit{back} & 48.67 & 24.93 & 65.29 & 67.26 & 48.3 & 50.89 \\
    \rowcolor[RGB]{231, 248, 254}w/ \textit{back} & 50.14 & 27.63 & 70.69 & 70.63 & 51.74 & 54.166 \\
     \multicolumn{1}{l}{\gray{$\Delta$ \textit{Gain}}} & \gbf{+1.47} & \gbf{+2.70} & \gbf{+5.40} & \gbf{+3.37} & \gbf{+3.44} & \gbf{+3.276} \\
     \bottomrule[1.2pt]
   \end{tabular}
\end{table*}

\begin{figure}[!t]
    \centering
    \includegraphics[width=0.99\columnwidth]{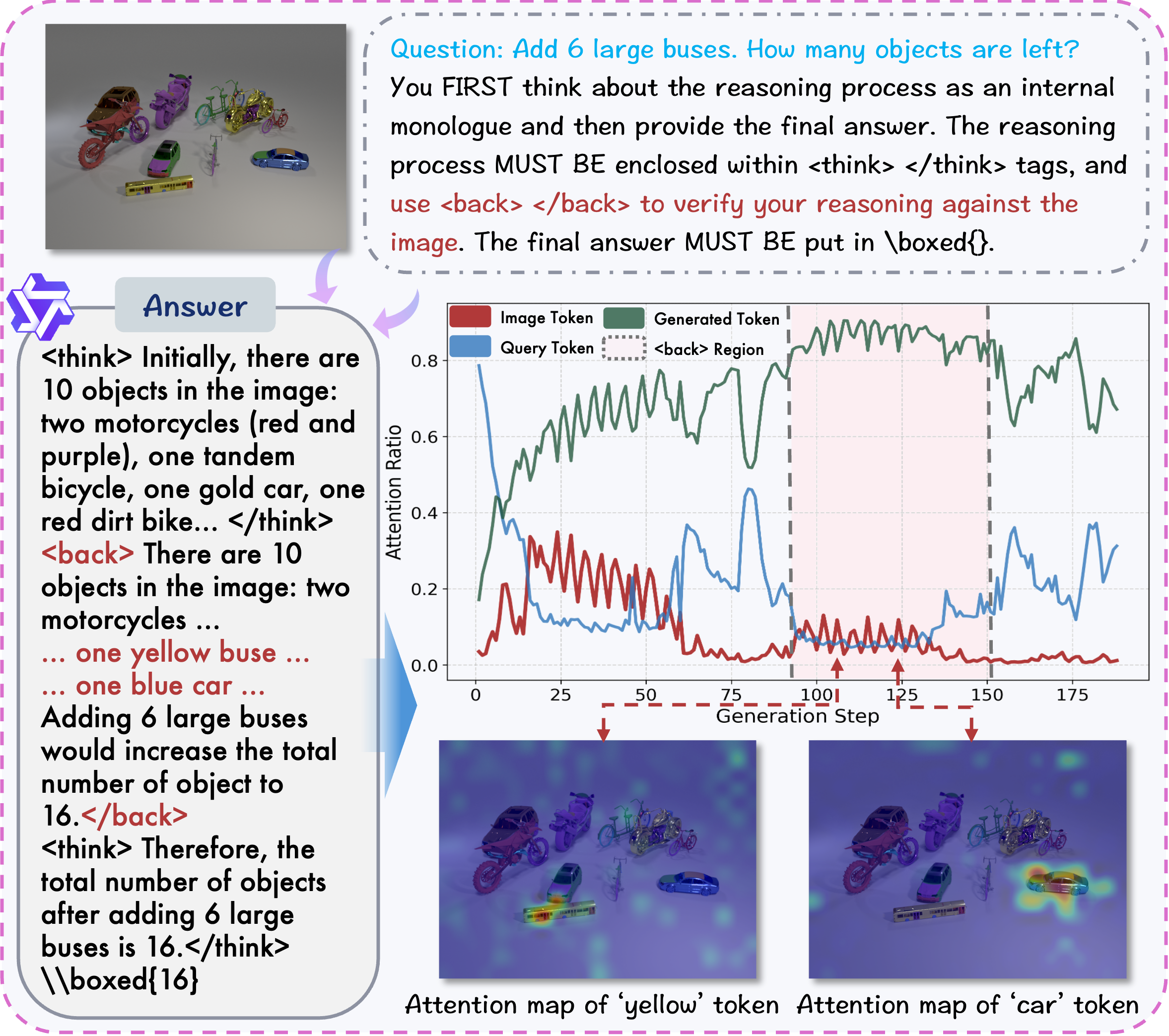}
    \caption{This figure shows how a modified prompt can encourage MLLM to spontaneously generate \texttt{<back>}  tokens and re-examine its reasoning against visual information. This triggers the model to autonomously re-focus its attention on specific visual details, like the yellow bus and car shown in the attention maps, to verify its conclusions.}
    \label{fig:2}
\end{figure}

However, this raises a fundamental question: \textbf{Can MLLMs spontaneously reactivate their attention to visual inputs without external intervention?} To investigate this, we conducted a preliminary experiment using a simple prompt modification that encourages the model to generate a \texttt{<back>} token and subsequently re-examine its response based on visual information.

Surprisingly, as shown in the Figure \ref{fig:2}, the model demonstrates a remarkable capacity for spontaneous visual attention recovery. Upon generating the \texttt{<back>}  token, the model naturally redirects substantial attention back to the visual input, evidenced by the sharp increase in the “Image Token” attention ratio shown in the central graph. Critically, this is not merely a general glance at the image; the model's reasoning becomes precisely grounded in the visual evidence. The attention maps on the bottom provide compelling proof: during the generation of the \texttt{<back>}  sequence, the model specifically focuses on the corresponding objects—for instance, attending to the yellow bus when generating the ``yellow" token and to the gold car for the ``car" token. This targeted refocusing occurs intrinsically, without any explicit re-injection of visual information or structural modifications to the model's architecture.

The results in Table~\ref{table:7b_results_initial_experiment} show quantitative improvements across multiple benchmarks, which initially validates that MLLMs possess latent capabilities for self-directed visual reflection. 
To further verify the performance gains brought by the back mechanism, we conducted a focused analysis specifically on the subset of questions where the ``Back prompt'' successfully triggered the visual reflection.
As detailed in Table~\ref{table:7b_results_with_back_gain}, comparing the performance on this specific subset of questions reveals that engaging in visual reflection leads to even greater improvements across all benchmarks.
However, the ``Trigger rate'' in Table~\ref{table:7b_results_initial_experiment} indicates a critical limitation: even with carefully tuned prompts, only modifying prompts is insufficient to consistently trigger this reflective behavior, resulting in an average trigger rate of only 62.48\%. Therefore, we propose using reinforcement learning to incentivize this mechanism further.

\section{Method of Look-Back}

The proposed \textbf{Look-Back} method is designed to guide multimodal large language models (MLLMs) to spontaneously re-focus visual inputs during inference, thereby enhancing their capability for visual fusion reasoning. Specifically, the Look-Back method comprises two primary stages: supervised fine-tuning (SFT) and reinforcement learning (RL).

\begin{figure*}[t]
    \centering
    \includegraphics[width=0.95\textwidth]{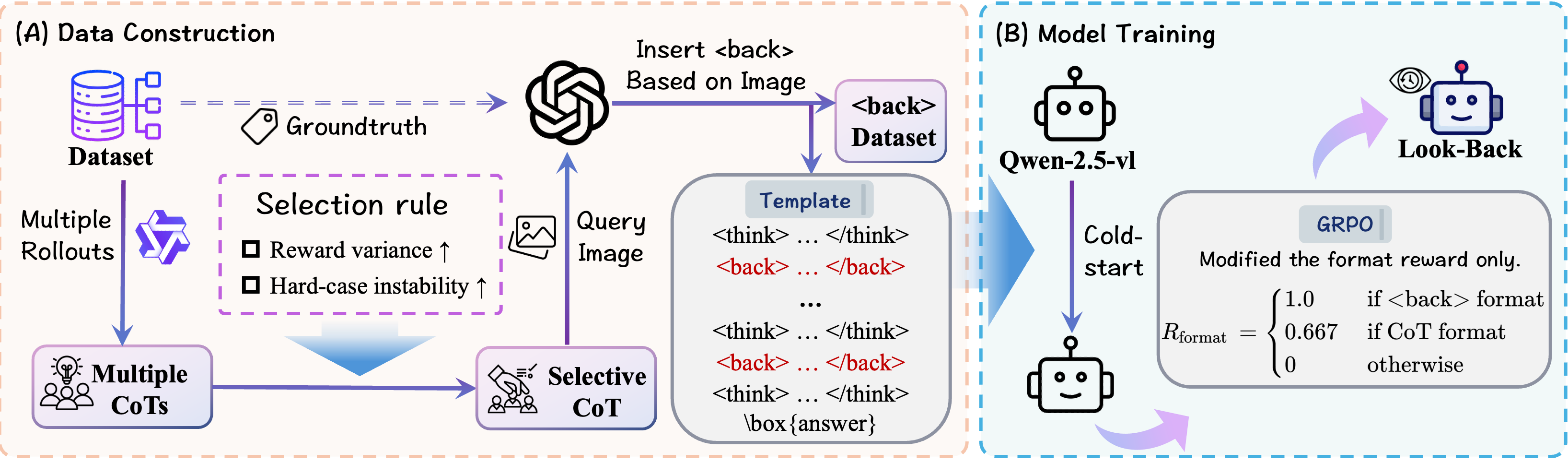}
    \caption{Pipeline of Look-Back Method, including Data Construction for Reflective SFT and Reinforcement Learning with Modified Rewards.}
    \label{fig_pipeline_2}
\end{figure*}

\subsection{Cold-start Initialization}

To address instability associated with the spontaneous triggering of the \texttt{<back>} token and reward hacking by the model (detailed in the Discussion), we first constructed a supervised fine-tuning dataset for cold-start initialization. Specifically, depending on when the \texttt{<back>} token is triggered, we classify the backtracking prompts into two categories:

\begin{itemize}
\item \textbf{Semantic-level backtracking (Semantic-back):} Triggered during the reasoning process, allowing the model to revisit visual details crucial for intermediate reasoning steps, and subsequently continue its ongoing reasoning.
\item \textbf{Solution-level backtracking (Solution-back):} Triggered after the model has generated a preliminary solution, prompting the model to rethink comprehensively by reconsidering visual input.
\end{itemize}
We designed two explicit output formats as follows (see Appendix~\ref{app:Prompt_Template} for complete details).

\begin{tcolorbox}[colback=gray!20!white, colbacktitle=white, coltitle=black, colframe=black!75!black, boxrule=0.7pt, halign title=center, title=\textbf{Output Formats of Backtracking}]
\textbf{Semantic-back:} \textless{}think\textgreater{} reasoning process here \textless{}/think\textgreater{} \textbf{\textcolor{red}{\textless{}back\textgreater{} verification process here \textless{}/back\textgreater{}}} \textless{}think\textgreater{} continue reasoning \textless{}/think\textgreater{} \textbackslash boxed\{final answer\}. \\

\noindent

\textbf{Solution-back:} \textless{}think\textgreater{} reasoning process here \textless{}/think\textgreater{} \textbf{\textcolor{red}{\textless{}back\textgreater{} verification process against the image here \textless{}/back\textgreater{}}} \textless{}think\textgreater{} based on the thinking  and verification contents, rethinking here \textless{}/think\textgreater{} \textbackslash boxed\{final answer\}.
\end{tcolorbox}

\textbf{Data Construction.} We designed a specific data construction process, as illustrated in Figure~\ref{fig_pipeline_2} (A), which consists of the following three steps:
\begin{enumerate}[nolistsep, leftmargin=*]
    \item \textbf{Model Inference:} First, we employ Qwen-2.5-VL-7B to perform Chain-of-Thought (CoT) inference on the dataset. For each question, we conduct $n$ independent inferences (with $n = 12$ in our experiments).
    \item \textbf{CoT Selection:} Based on the inference results, we calculate the accuracy reward and select the questions that have a higher reward variance and greater difficulty.
    \item \textbf{Advanced Model Insertion:} The question, image, model-generated CoT reasoning process, and the ground-truth answer are input into GPT-o4-mini, which automatically inserts the backtracking tokens based on predefined rules. For samples with correct answers, backtracking tokens related to image validation are inserted. For samples with incorrect answers, backtracking tokens that correct the answer based on image information are inserted, and the final answer is adjusted accordingly.
\end{enumerate}

Through the above steps, each sample receives a stable cold-start response with clearly marked tokens. This yields a stable cold-start dataset with explicit backtracking markers.

\textbf{Supervised Fine-Tuning (SFT).} Using the cold-start dataset generated with the \texttt{<back>} tokens, we apply SFT to guide the model to consistently trigger the backtracking behavior.  Each sample is represented as $(x, q, r_\mathrm{back}, a)$, where $x$ denotes the input image, $q$ represents the question, $r_\mathrm{back}$ is the backtracking token sequence, and $a$ is the answer sequence. The training objective is as follows:
\begin{equation}
    \mathcal{L}_{\mathrm{cold-start}} = -\mathbb{E}_{(x, q, r_\mathrm{back}, a) \sim \mathcal{D}} \sum_{t=1}^{|y|} \log \pi_{\theta}\left(y_t \mid x, q, y_{<t}\right),
\end{equation}
where $\mathcal{D}$ denotes the dataset, and $y = [r_\mathrm{back}; a]$ concatenates the backtracking tokens and answer sequence.

\subsection{Look-Back Reinforcement Learning}
To further enhance the model's ability to autonomously revisit visual inputs, we employed the Group Relative Policy Optimization (GRPO) algorithm for reinforcement learning. Compared to traditional policy optimization methods, GRPO performs policy gradient optimization within a sample group, enabling the model to generate more diverse and rich reasoning responses efficiently. The optimization objective is as follows:
\begin{equation}
  \resizebox{\linewidth}{!}{$%
    \begin{aligned}
      \mathcal{J}_{\text{GRPO}}(\theta)
      & =\mathbb{E}\!\left[q \sim P(Q),\{o_i\}_{i=1}^{G} \sim \pi_{\theta_{\text{old}}}(O \mid q)\right] \\
      &\quad \frac{1}{G}\sum_{i=1}^{G}\frac{1}{|o_i|}\sum_{t=1}^{|o_i|}\!\Bigl\{
          \min\!\Bigl[
            \frac{\pi_{\theta}(o_{i,t}\!\mid\! q, o_{i,<t})}{\pi_{\theta_{\text{old}}}(o_{i,t}\!\mid\! q, o_{i,<t})}
            A_{i,t}, \\
      & 
            \operatorname{clip}\!\Bigl(
              \frac{\pi_{\theta}(o_{i,t}\!\mid\! q, o_{i,<t})}{\pi_{\theta_{\text{old}}}(o_{i,t}\!\mid\! q, o_{i,<t})},
              1-\varepsilon, 1+\varepsilon
            \Bigr) A_{i,t}
          \Bigr] -\beta\,\mathbb{D}_{\mathrm{KL}}\!\bigl[\pi_{\theta}\,\|\,\pi_{\mathrm{ref}}\bigr]\Bigr\},
    \end{aligned}$}%
\end{equation}
where $\epsilon$ and $\beta$ are the clipping hyperparameters and the KL divergence penalty coefficient, respectively. To guide the model in triggering the visual review behavior more stably, we modified only the format reward function. Specifically, the format reward function $R_{\text{format}}$ is defined as:
\begin{equation}
    R_{\text{format}} =
    \begin{cases}
    1.0, & \text{if <back> format}, \\
    0.667, & \text{if CoT format}, \\
    0, & \text{otherwise}.
    \end{cases}
\end{equation}

The complete reward function is a combination of the format reward and accuracy reward, defined as:
\begin{equation}
R = \lambda \cdot R_{\text{format}} + R_{\text{accuracy}},
\end{equation}
where $R_{\text{accuracy}}$ represents the accuracy reward for the response, and $\lambda$ is a hyperparameter used to adjust the balance between the format reward and the accuracy reward. Essentially, the reward function we designed provides the model with an intrinsic motivation to autonomously revisit visual information. This enables the model to actively reflect on visual inputs during the reasoning process, similar to how humans naturally revisit visual information, without the need for explicit re-injection of images.

\section{Look-Back Experiments Analysis}

\subsection{Experimental Setup}

\textbf{Baselines and Benchmarks.}
To evaluate the effectiveness of Look-Back, we conducted experiments on a set of eight benchmarks, divided into two categories: mathematical and perceptual tasks. The mathematical benchmarks include MathVerse~\cite{zhang2024mathverse}, MathVision~\cite{wang2024measuring}, MathVista~\cite{lu2023mathvista}, WeMath~\cite{qiao2024we}, and GeoMath~\cite{tan2025reason}, while the perceptual benchmarks consist of HallusionBench~\cite{guan2024hallusionbench}, TallyQA~\cite{acharya2019tallyqa}, and MME~\cite{fu2024mmecomprehensiveevaluationbenchmark}. We computed the average performance for each category separately. Additionally, we compared Look-Back against three types of baselines: (1) Closed-Source Multimodal Large Language Models (MLLMs), such as GPT-4o~\cite{hurst2024gpt} and o3~\cite{openai2025o3o4systemcard}; (2) Open-Source General MLLMs, , such as Qwen2.5-VL-32B~\cite{qwen2.5-VL} and InternVL3-38B~\cite{zhu2025internvl3}; and (3) Open-Source Reasoning MLLMs, such as MM-Eureka-8B~\cite{meng2025mmeureka}, R1-VL-7B~\cite{zhang2025r1}, VL-Rethinker-7B~\cite{wang2025vl}, OpenVLThinker-7B~\cite{deng2025openvlthinker}, ThinkLite-VL-7B~\cite{wang2025sota}, VLAA-Thinker-7B~\cite{chen2025sft}, Vision-R1-7B~\cite{huang2025vision}, MM-Eureka-Qwen-7B~\cite{meng2025mmeureka}, R1-Onevision-7B~\cite{yang2025r1}, and NoisyRollout-7B~\cite{liu2025noisyrollout}.

\textbf{Training Datasets.}
For the reinforcement learning (RL) phase, we selected 15k mathematical problems from the Geo170K~\cite{gao2023g}, Math360K~\cite{shi2024math}, Geometry3K~\cite{lu2021inter}, and K12~\cite{meng2025mmeureka} datasets for training. During the supervised fine-tuning (SFT) phase, we applied the data construction process outlined in Section 3.1 to the 15k problems from the RL phase, generating 4k and 10k cold-start datasets for Semantic-back and Solution-back, respectively.

\textbf{Implementation Details.}
The training was conducted on eight NVIDIA A800 GPUs, where we performed cold-start SFT and subsequent RL training on the Qwen2.5-VL-7B-Instruct model. We used the LLaMA-Factory~\cite{zheng2024llamafactory} framework for SFT. To prevent overfitting, we trained for only one epoch. For RL, we employed the EasyR1~\cite{sheng2024hybridflow, zheng2025easyr1} framework, where the default reward weight, denoted by $\lambda$, was set to 0.1. Training was carried out for two epochs on the 15k dataset, using a batch size of 128 (with 12 rollouts per sample) and a sampling temperature of 1.0. Additional settings can be found in Appendix~\ref{app:Experimental details}.

\begin{table*}[t]
  \centering
  \caption{Comparison of our Look-Back models (\textit{Semantic-back} and \textit{Solution-back}) with representative \textbf{Closed-Source}, \textbf{Open-Source General}, and \textbf{Open-Source Reasoning MLLMs} across the \textbf{Math-Benchmark} and \textbf{Perception-Benchmark} suites (higher is better). $^\dagger$\,scores are taken from the respective models’ official reports.}
  \label{table:all}

  \resizebox{\textwidth}{!}{%
  \setlength{\tabcolsep}{3.4pt}
  \begin{tabular}{lcccccc|cccc|c}
    \toprule[1.2pt]
    \textbf{Model} &
    \multicolumn{6}{c}{\textbf{Math‐Benchmark}} &
    \multicolumn{4}{c}{\textbf{Perception‐Benchmark}} &
    \textbf{Overall} \\
    \cmidrule(lr){2-7} \cmidrule(lr){8-11}
     & MathVerse & MathVision & MathVista & WeMath & GeoMath & Avg\textsubscript{M} &
       Hallusion &  TallyQA & MME & Avg\textsubscript{P} &
       Avg\textsubscript{All} \\
    \midrule

    \multicolumn{12}{c}{\textbf{Closed-Source MLLMs}}\\
    \midrule
    Claude-3.7 & 52$^\dagger$  & 41.3$^\dagger$  & 66.8$^\dagger$  & 72.6$^\dagger$  & -  & -  & -  & -  & -  & -  & -  \\
    GPT-4o & 50.8$^\dagger$  & 30.4$^\dagger$  & 63.8$^\dagger$  & 69$^\dagger$  & -  & -  & -  & -  & -  & -  & - \\
    GPT-o1 & 57$^\dagger$  & 60.3$^\dagger$  & 73.9$^\dagger$  & 98.7$^\dagger$  & -  & -  & -  & -  & -  & -  & -  \\
    GPT-o3 & -  & -  & 86.8$^\dagger$  & -  & -  & -  & -  & -  & -  & -  & -  \\
    Gemini-2-flash & 59.3$^\dagger$  & 41.3$^\dagger$  & 70.4$^\dagger$  & 71.4$^\dagger$  & -  & -  & -  & -  & -  & -  & -  \\
    \midrule
    \multicolumn{12}{c}{\textbf{Open-Source General MLLMs (7B-38B)}}\\
    \midrule
    InternVL2.5-8B & 39.5$^\dagger$  & 19.7$^\dagger$  & 64.4$^\dagger$  & 53.5$^\dagger$  & 63  & 48.0  & 61.7  & 53.9  & -  & -  & -  \\
    InternVL2.5-38B & 49.4$^\dagger$  & 31.8$^\dagger$  & 71.9$^\dagger$  & 67.5$^\dagger$  & -  & -  & 70.0  & -  & -  & -  & -    \\
    InternVL3-8B & 39.8$^\dagger$  & 29.3$^\dagger$  & 71.6$^\dagger$  & -  & 45.6  & -  & 64.3  & -  & 85.1 / 2322  & -  & -  \\
    InternVL3-38B & 48.2$^\dagger$  & 34.2$^\dagger$  & 75.1$^\dagger$  & -  & 48.2  & -  & 72.0  & 75.1  & 87.7 / 2403  & 78.3  & -    \\
    QwenVL2.5-7B & 46.3$^\dagger$  & 25.1$^\dagger$  & 68.2$^\dagger$  & 62.1$^\dagger$  & 45.6  & 49.5  & 65.0  & 75.5  & 82.1 / 2180  & 74.2  & 61.8 \\ 
    QwenVL2.5-32B & 48.5$^\dagger$  & 38.4$^\dagger$  & 74.7$^\dagger$  & 69.1$^\dagger$  & 54.5  & 57.0  & 71.8  & 79.2  & 88.4 / 2444 & 79.8  & 68.4  \\ 
    \midrule

    \multicolumn{12}{c}{\textbf{Open‐Source Reasoning MLLMs (7B)}} \\
    \midrule
    MM-Eureka-8B & 40.4$^\dagger$  & 22.2$^\dagger$  & 67.1$^\dagger$  & 58.7  & 50.7  & 47.8  & 65.3  & 76.9  & 84.4 / 2306  & 75.5  & 61.7 \\ 
    R1-VL-7B & 40.0$^\dagger$  & 24.7$^\dagger$  & 63.5$^\dagger$  & 60.1  & 47.7  & 47.2  & 54.7  & 72.9  & 86.4 / 2376$^\dagger$  & 71.3  & 59.3  \\ 
    VL-Rethinker-7B & 52.9  & 30.0  & 74.4  & 69.1  & 50.0  & 55.3  & 69.9  & 76.5  & 86.9 / 2336  & 77.8  & 66.0  \\ 
    OpenVLThinker-7B & 45.7  & 26.3  & 71.2  & 66.7  & 55.0  & 53.0  & 70.2  & 80.1  & 86.4 / 2328  & 78.9  & 65.5  \\ 
    ThinkLite-VL-7B & 49.3  & 26.2  & 71.7  & 61.9  & 46.5  & 51.1  & 70.7  & 80.3  & 87.6 / 2378  & 79.5  & 65.6  \\ 
    VLAA-Thinker-7B & 52.7  & 29.2  & 69.7  & 70.2  & 48.8  & 54.1  & 68.2  & 78.2  & 84.8 / 2356  & 77.1  & 65.3  \\
    Vision-R1-7B & 52.4$^\dagger$  & 28.0  & 70.6  & 73.9  & 48.6  & 54.7  & 65.5  & 78.1  & 84.4 / 2312  & 76.0  & 65.3  \\ 
    MM-Eureka-Qwen-7B & 50.5  & 28.9  & 70.4  & 65.2  & 47.7  & 52.5  & 68.6  & 78.3  & 86.1 / 2370  & 77.7  & 65.1  \\ 
    R1-Onevision-7B & 46.4  & 29.9  & 64.1  & 61.8  & 47.7  & 50.0  & 67.5  & 76.7  & 82.3 / 2284  & 75.5  & 62.7  \\ 
    NoisyRollout-7B & 53.2$^\dagger$  & 27.8  & 72.5$^\dagger$  & 70.8  & 50.7  & 55.0  & 70.8  & 77.4  & 81.8 / 2038 & 76.7  & 65.8 \\ 
    \midrule
    \rowcolor[RGB]{231, 248, 254}Semantic-back-7B & 50.5  & 27.7  & 71.6  & 71.3  & 56.5  & 55.5  & 70.7  & 81.2  & 87.1 / 2340 & 79.6  & 67.6  \\ 
    \rowcolor[RGB]{231, 248, 254}Solution-back-7B & 51.8  & 30.3  & 72.3  & 70.8  & 56.7  & 56.4  & 69.8  & 79.2  & 85.9 / 2319 & 78.3  & 67.3  \\ 
    \bottomrule[1.2pt]
  \end{tabular}}
\end{table*}

\begin{table*}[!ht]
  \centering
  \caption{Ablation of the \textit{Look-Back},  selectively removing SFT or RL for both \textit{Semantic-back} and \textit{Solution-back}. }
  \label{table:Ablation}

  \resizebox{\textwidth}{!}{%
  \setlength{\tabcolsep}{4.5pt}
  \begin{tabular}{lcccccc|cccc|c}
    \toprule[1.2pt]
    \textbf{Model} &
    \multicolumn{6}{c}{\textbf{Math‐Benchmark}} &
    \multicolumn{4}{c}{\textbf{Perception‐Benchmark}} &
    \textbf{Overall} \\
    \cmidrule(lr){2-7} \cmidrule(lr){8-11}
     & MathVerse & MathVision & MathVista & WeMath & GeoMath & Avg\textsubscript{M} &
       HallusionBench &  TallyQA & MME & Avg\textsubscript{P} &
       Avg\textsubscript{All} \\
    \midrule
    Qwen-2.5-VL-7B & 46.3$^\dagger$  & 25.1$^\dagger$  & 68.2$^\dagger$  & 62.1$^\dagger$  & 45.6  & 49.5  & 65.0  & 75.5  & 82.1  & 74.2  & 61.8 \\ 
    +GRPO & 49.3  & 26.8  & 70.9  & 67.6  & 55.2  & 53.9  & 68.6  & 78.3  & 85.5  & 77.5  & 65.7  \\ 
    \midrule
    \rowcolor[RGB]{231, 248, 254} Semantic-back-7B & 50.5  & 27.7  & 71.6  & 71.3  & 56.5  & 55.5  & 70.7  & 81.2  & 87.1  & 79.6  & 67.6  \\ 
    w/o \textit{SFT} & 49.7  &  27.3 & 71.3  & 70.1  & 56.3  & 54.9  & 69.3  & 79.5 & 86.6  & 78.5  & 66.7    \\ 
    w/o \textit{RL} & 44.7  & 24.4  & 63.8  & 58.9  & 37.9  & 45.9  & 68.5  & 67.4  & 77.1  & 71.0  & 58.5  \\ 
    \midrule
    \rowcolor[RGB]{231, 248, 254} Solution-back-7B & 51.8  & 30.3  & 72.3  & 70.8  & 56.7  & 56.4  & 69.8  & 79.2  & 85.9  & 78.3  & 67.3  \\ 
    w/o \textit{SFT} & 49.5  & 27.9  & 72.0  & 70.3  & 56.2  & 55.2  & 69.3  & 79.1  & 86.0  & 78.1  & 66.7  \\ 
    w/o \textit{RL} & 43.4  & 20.2  & 63.1  & 52.3  & 36.2  & 43.0  & 65.0  & 74.3  & 83.4  & 74.2  & 58.6  \\ 
    \bottomrule[1.2pt]
  \end{tabular}}
\end{table*}

\subsection{Main Results}

\textbf{Mathematical Reasoning.}
As shown in Table~\ref{table:all}, our Look-Back approach, built on the Qwen2.5-VL-7B, outperforms the base model across all benchmarks. Specifically, on five mathematical benchmarks, Semantic-back improved by an average of 7\% (from 48.5\% to 55.5\%), and Solution-back showed an enhancement of 7.9\% (from 48.5\% to 56.4\%). Furthermore, we compared Look-Back with ten different Open-Source Reasoning MLLMs. Although the training data and duration varied across models, making a direct comparison challenging, Look-Back still demonstrated competitive performance. Despite having significantly fewer parameters, Solution-back narrowed the gap with closed-source models, thanks to the ``look-back" mechanism.

\textbf{Perceptual Reasoning.}
Although our training primarily utilized mathematical reasoning data, it is noteworthy that on the perceptual benchmarks, Semantic-back achieved an average improvement of 6.3\% (from 61.3\% to 67.6\%) and Solution-back showed a 6\% increase (from 61.3\% to 67.3\%) compared to the baseline model. Additionally, our approach exhibited strong competitiveness with other Open-Source Reasoning MLLMs. These results underscore the significance of the ``look-back" mechanism in enhancing the generalization capabilities of multimodal reasoning systems.

\subsection{Ablation Study}
\textbf{Effectiveness of Look-Back.}
We further investigate the contributions of each stage within the Look-Back framework. As shown in Table~\ref{table:Ablation}, removing either the RL or SFT phase of the look-back training leads to a significant degradation in model performance. Moreover, when compared to the standard GRPO without any look-back mechanism, both the Semantic-level and Solution-level back mechanisms demonstrate performance improvements through the application of look-back. Further analysis of the training process can be found in Appendix~\ref{app: Training_Dynamics}.

\textbf{Ablation of Reflection Rate.}
Since the model's look-back process consists of both verification and reflection-based error correction, it is unreasonable to provide a single look-back dataset during the SFT cold-start phase, as this would easily lead to reward hacking. Therefore, we conducted an ablation study on the reflection rate of the SFT dataset, using the Semantic-level back mechanism as an example. The results, illustrated in Table~\ref{table:RR}, show that the optimal reflection rate for different types of tasks lies between 30\% and 50\%. Both excessively low and high reflection rates result in a decrease in model performance. As a result, we adopted a reflection rate of 50\% in this study.

\subsection{Reasoning Qualitative Analysis}

\begin{figure*}[!t]
    \centering
    \includegraphics[width=0.95\textwidth]{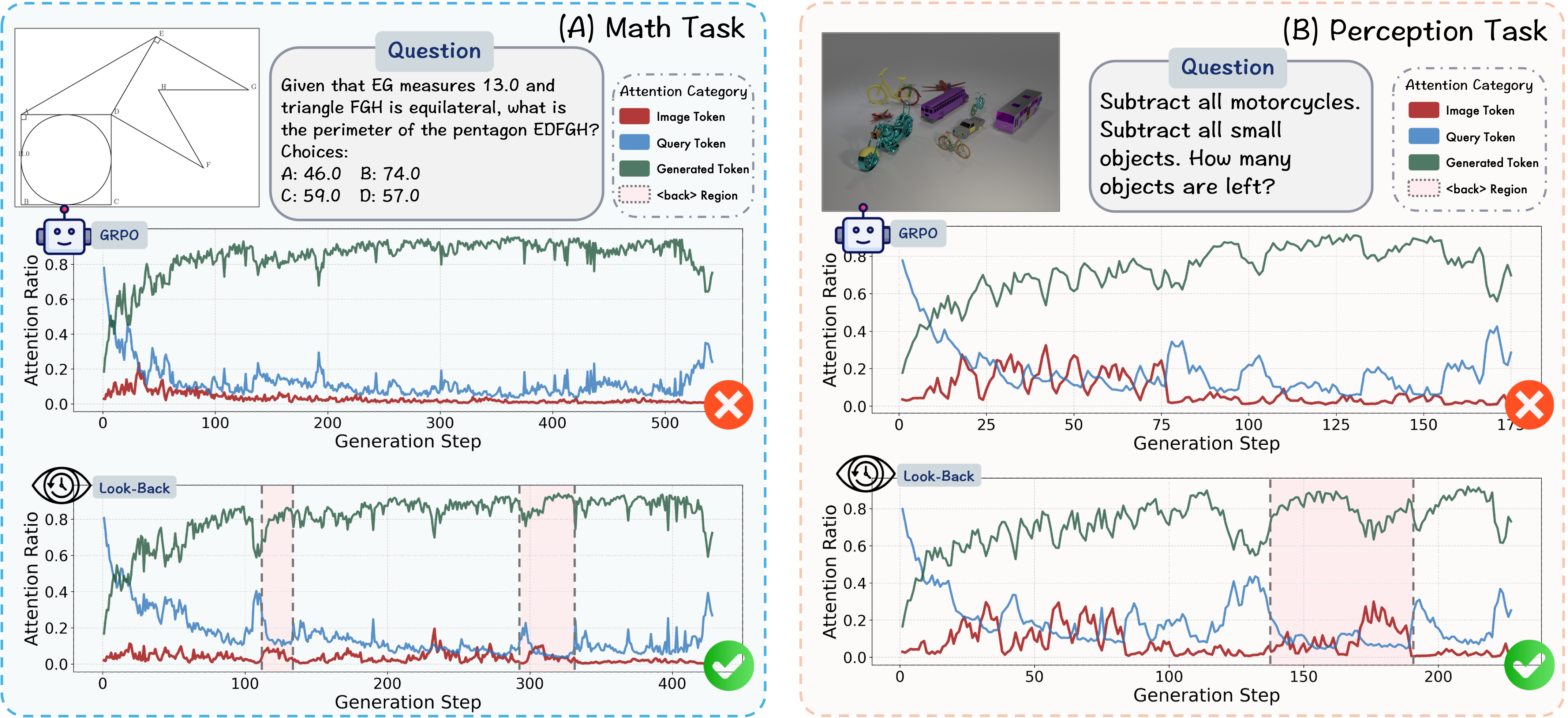}
    \caption{Look-Back Enhances Visual Grounding with Multiple Verifications. The graphs illustrate that, unlike models trained with standard GRPO, our model successfully and repeatedly re-focuses on visual input (spikes in red line) during the later reasoning stages for both Math (A) and Perception (B) tasks. This visual verification can occur multiple times within one task, demonstrating an autonomous ability to revisit and ground reasoning in visual evidence.}
    \label{fig:case}
\end{figure*}

\begin{table*}[t]
  \centering
  \caption{Effect of reflection-rate on performance. \textit{RR-$x$\%} denotes training with an $x\%$ reflection rate.}
  \label{table:RR}

  \resizebox{\textwidth}{!}{%
  \setlength{\tabcolsep}{3.4pt}
  \begin{tabular}{lcccccc|cccc|c}
    \toprule[1.2pt]
    \textbf{Model} &
    \multicolumn{6}{c}{\textbf{Math‐Benchmark}} &
    \multicolumn{4}{c}{\textbf{Perception‐Benchmark}} &
    \textbf{Overall} \\
    \cmidrule(lr){2-7} \cmidrule(lr){8-11}
     & MathVerse & MathVision & MathVista & WeMath & GeoMath & Avg\textsubscript{M} &
       HallusionBench &  TallyQA & MME & Avg\textsubscript{P} &
       Avg\textsubscript{All} \\
    \midrule
    RR-10\% & 49.0  & 26.6  & 69.9  & 68.1  & 56.7  & 54.1  & 68.8  & 80.5  & 86.7  & 78.6  & 66.3  \\ 
    RR-30\% & 50.8  & 28.8  & 71.5  & 69.9  & 56.8  & 55.6  & 69.5  & 80.3  & 86.1  & 78.6  & 67.1  \\
    \rowcolor[RGB]{231, 248, 254}RR-50\% & 50.5  & 27.7  & 71.6  & 71.3  & 56.5  & 55.5  & 70.7  & 81.2  & 87.1  & 79.6  & 67.6  \\
    RR-70\% & 50.0  & 27.6  & 70.6  & 69.1  & 55.0  & 54.5  & 68.8  & 79.7  & 86.3  & 78.2  & 66.4  \\
    RR-90\% & 49.9  & 27.2  & 70.7  & 70.0  & 56.6  & 54.9  & 70.4  & 80.6  & 86.9  & 79.3  & 67.1  \\
    \bottomrule[1.2pt]
  \end{tabular}}
\end{table*}
Beyond the quantitative performance improvements observed across various benchmarks, we conducted qualitative analyses to verify that Look-Back alters MLLM attention patterns. Specifically, As shown in Figure~\ref{fig:case}, our method consistently improves attention across both mathematical and perceptual tasks. 
Compared to standard GRPO, Look-Back enables models to re-focus on visual input during later reasoning stages for verification.

Further qualitative analyses (Appendix~\ref{app: Case_study}) reveal concrete cases from five different benchmarks, highlighting how both \textit{Semantic-back} and \textit{Solution-back} effectively utilize the Look-Back mechanism to rectify initial errors by explicitly grounding reasoning in visual evidence.
This demonstrates that Look-Back effectively guides MLLMs to autonomously determine when, where, and how to revisit visual information, thereby moving beyond sole reliance on text-based reasoning. 
This finding further supports our key insight: with proper guidance, MLLMs can perform visual fusion reasoning without explicit visual prompting.

\section{Further Discussion}

\subsection{Failed Attempts}
During our attempts to leverage the model's ability to spontaneously re-focus on images, we encountered several failures and setbacks. In this section, we analyze these failed experiences, though we emphasize that such failures do not imply the approach itself was fundamentally flawed.

\textbf{Reward Hacking in Weaker Models.}
We initially applied Look-Back training on the Qwen-2-VL model but encountered reward hacking: the model learned a shortcut by generating an empty \texttt{<back></back>} token sequence, thus obtaining format rewards without genuine reasoning. This aligns with prior findings~\cite{yue2025does} that reinforcement learning may fail to enhance reasoning beyond the base model. We hypothesize that this issue arises because Qwen-2-VL inherently lacks sufficient capability for visual reflection, whereas Qwen-2.5-VL may possess this ability due to its pretraining.

\textbf{SFT Cold-Start Data Requirements.}
Initially, we generated CoT data using GPT-4o and subsequently inserted the \texttt{<back>} tokens. However, we observed a deterioration in performance after cold-starting the model.  Inspired by \citet{wan2025srpo}, we instead used model-generated data with refined \texttt{<back>} insertion, resulting in improved performance. We hypothesize that fine-tuning on homologous model outputs reduces distributional deviation, aligning better with the cold-start objective of consistent output formatting.

\subsection{Impact of Cold Start}

\begin{figure}[!t]
    \centering 
    \includegraphics[width=0.95\columnwidth]{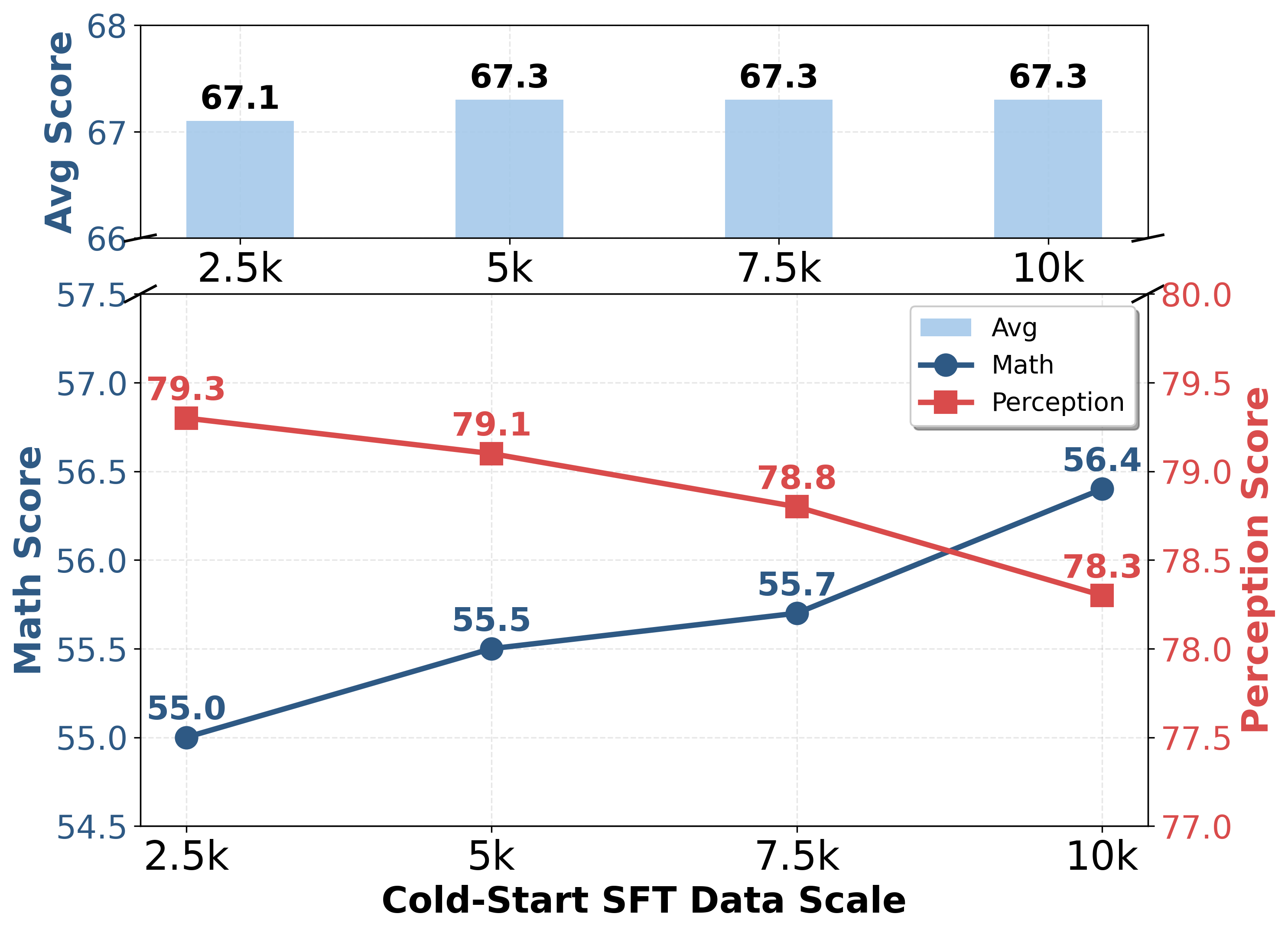} 
    \caption{Impact of cold-start SFT scale (2.5k → 10k) on model performance: Math scores rise steadily, Perception scores decline marginally, and the overall average remains almost flat.}
    \label{fig:scale} 
\end{figure}
\textbf{Scaling Cold-Start Data.}
To assess the effect of cold-start data scale on performance, we experimented with 2.5k, 5k, 7.5k, and 10k samples, all mathematical in nature, using the Solution-back method. As shown in Figure~\ref{fig:scale}, with an increase in cold start data, the average score for mathematical tasks improved, demonstrating that scaling during the cold start phase contributes to continuous performance improvement. However, performance on perceptual tasks declined slightly, although the overall performance remained relatively unchanged. We hypothesize that cold starting with purely mathematical data may limit further generalization on perceptual tasks. Incorporating more diverse SFT and RL data could further enhance overall robustness.

\textbf{Performance Differences Between Semantic-Back and Solution-Back.} As illustrated in Table~\ref{table:Ablation}, triggering both types of \texttt{<back>} methods enhances performance on multiple benchmarks. Semantic-back performs better on perceptual tasks, while Solution-back excels on mathematical ones. 
We speculate that early backtracking facilitates timely confirmation of visual cues, benefiting perceptual tasks. In contrast, deferring backtracking until after CoT reasoning enables more comprehensive verification with minimal disruption to the reasoning chain, favoring mathematical tasks.

\section{Related Work}

Multimodal complex reasoning has advanced significantly in recent years, evolving through four main stages: early explicit module exploration, supervised fine-tuning and testing-time scaling, reinforcement learning-driven advancements, and the continued evolution of multimodal alignment and native visual reasoning capabilities.

\textbf{Early Development of Multimodal Reasoning}~\cite{shao2024visual, zhang2023multimodal, hu2024visual}.
In the early stages of MLLM development, multimodal reasoning relied on explicit prompts and multi-module cooperation. Techniques like Visual-CoT~\cite{shao2024visual} used reasoning chains and visual sampling for dynamic visual reasoning. Visual-SketchPad~\cite{hu2024visual} introduced a three-stage workflow incorporating visual sketching to enhance interpretability. Meanwhile, Multimodal-CoT~\cite{zhang2023multimodal} proposed a two-stage framework that decouples reasoning chain generation from answer inference.

\textbf{Supervised Fine-Tuning and Test-Time Scaling}~\cite{xu2024llava, wang2025multimodal, du2025virgo, ma2024taco, yang2025asft, kumar2025llm, yang2024parameter}. With the emergence of models such as OpenAI O1~\cite{jaech2024openai}, supervised fine-tuning (SFT) based on large-scale synthetic chain-of-thought data became mainstream. The core feature of this paradigm shift was the transition from module-based methods to data-driven approaches. For example, Virgo~\cite{du2025virgo} dynamically adjusts the depth of reasoning by utilizing chain-of-thought data of varying lengths. LLaVA-CoT~\cite{xu2024llava} employs a structured reasoning template that constrains the model to follow a multi-step reasoning process. TACO~\cite{ma2024taco} applies dynamic programming strategies for tool invocation learning through SFT data. Test-Time Scaling (TTS)~\cite{ma2024taco, kumar2025llm, muennighoff2025s1, zhang2023multimodal} further enhances reasoning without updating model parameters, establishing a foundation for reinforcement learning methods.

\textbf{Reinforcement Learning Breakthroughs}~\cite{lightman2023let, wang2025vl, meng2025mmeureka, zhang2025r1, park2025dip, yu2025perception, li2025star, liu2025infi, wang2025jigsaw, yu2025dapo, feng2025video, liu2025seg, zhou2025r1, wang2025crowdvlm, liu2025x, xia2025visionary, yao2025r1, ma2025one}. The success of DeepSeek-R1~\cite{guo2025deepseek} marked the entry of complex reasoning into a new era of reinforcement learning fine-tuning (RFT). In the multimodal domain, DIP-R1~\cite{park2025dip} explored fine-grained image processing, while Perception-R1~\cite{yu2025perception} encoded image patches directly, effectively integrating testing-time augmentation methods with RFT training. MM-Eureka~\cite{meng2025mmeureka} made significant strides in visual reasoning through rule-based rewards. STAR-R1~\cite{li2025star}, VL-Rethinker~\cite{wang2025vl}, and Infi-MMR~\cite{liu2025infi} further demonstrated the effectiveness of reinforcement learning in spatial, medical~\cite{chen2024huatuogpt}, and embodied~\cite{zhang2025embodied, zhao2025embodied, shen2025phyx} reasoning.

\textbf{Evolution of Visual Thinking}~\cite{wu2024v, li2025imagine, li2025dyfo, feng2025reflective, zheng2025deepeyes, su2025openthinkimg, zhang2025chain, wang2025visuothink, chern2025thinking, wu2025lanp,sarch2025grounded, wu2025reinforcing, xu2025visual,chen2025ict, zhang2025mllms, gupta2023visual, chung2025don, zhao2025cot, wang2025visuothink, fu2025refocus, shen2024zoomeye}. Recent research trends indicate that multimodal complex reasoning not only requires ``thinking in language" but also necessitates ``thinking in images."~\cite{zheng2025deepeyes, sarch2025grounded, su2025openthinkimg, zhang2025chain, wang2025visuothink, chern2025thinking, wu2025reinforcing, zeng2025futuresightdrive, wang2025vgr} In the area of fine-grained perception, Vstar~\cite{wu2024v} introduced the SEAL framework, which dynamically locates key details through a hierarchical visual search mechanism. DyFo~\cite{li2025dyfo} simulates the dynamic focusing mechanism of human visual search, and DeepEyes~\cite{zheng2025deepeyes} achieves dynamic interplay between visual and textual reasoning through end-to-end reinforcement learning. In terms of complex spatial reasoning, MVoT~\cite{li2025imagine} alternates between generating text and images during the reasoning process, supplementing linguistic reasoning with visual thought processes. Reflective Planning~\cite{feng2025reflective} uses diffusion models to predict future visual states, creating a ``predict-reflect-correct" feedback loop.

Unlike previous methods that explicitly inject visual information~\cite{zheng2025deepeyes, su2025openthinkimg, zhang2025chain, wang2025visuothink, chern2025thinking, sarch2025grounded, wu2025reinforcing, xu2025visual, zhang2025mllms, gupta2023visual}, the Look-Back method enables models to autonomously learn when and how to refocus on visual input, enhancing reasoning capabilities without explicit visual guidance.

\section{Conclusion}
In this work, we observed that Multimodal Large Language Models (MLLMs) can autonomously re-focus their attention on visual inputs during reasoning, without explicit visual information injection. Building on this insight, we introduced the Look-Back approach, which empowers MLLMs to self-direct visual reflection through a two-stage training process combining supervised fine-tuning and reinforcement learning. Our experiments show that Look-Back significantly enhances multimodal reasoning capabilities, achieving competitive results across multiple benchmarks.

\section{Acknowledgment}
We thank Guowei Xu, Peng Jin, and Zhongwei Wan for their support in technical discussions related to this work. We also thank Yuyang Liu for providing computational resources.

\newpage
\bibliography{aaai2026}

\newpage

\appendix

\section{Experimental details}
\label{app:Experimental details}

\subsection{Dataset}
We construct our training dataset by selectively sampling from four established multimodal reasoning datasets, each featuring distinct geometric and mathematical reasoning characteristics:
\begin{itemize}[leftmargin=*]
    \item \textbf{Geo170K}~\citep{gao2023g}: A large-scale multimodal geometric dataset containing over 170K geometric image-text pairs, constructed through systematic data generation using existing datasets and text generation models. The dataset provides diverse geometric reasoning scenarios requiring integration of visual and textual information.
    
    \item \textbf{MathV360K}~\citep{shi2024math}: A comprehensive mathematical reasoning dataset comprising 360K multimodal mathematical problems, created by collecting high-quality image-question pairs and synthesizing additional samples. It systematically bootstraps multimodal reasoning by pairing mathematical prompts with visual contexts.
    
    \item \textbf{Geometry3K}~\citep{lu2021inter}: Contains 3K multiple-choice geometry problems with dense formal language annotations, including annotated diagram logical forms and textual descriptions. Over 99\% of problems require combining image information for correct solutions.
    
    \item \textbf{K12}~\citep{meng2025mmeureka}: An educational dataset covering mathematical concepts from elementary through high school levels, containing curated problems spanning various grade levels and mathematical domains.
\end{itemize}
From these datasets, we sample 15K mathematical problems for reinforcement learning training. During the supervised fine-tuning phase, we apply the data construction process outlined in Section 3.1 to generate 4K Semantic-back and 10K Solution-back cold-start samples, providing stable initialization for the Look-Back mechanism.

\subsection{Hyper-parameters}

\textbf{Supervised Fine-tuning (SFT) Stage:} We first performed cold-start supervised fine-tuning on the Qwen2.5-VL-7B-Instruct model. The training employed the LLaMA-Factory framework, utilizing a full parameter fine-tuning strategy rather than parameter-efficient fine-tuning methods. To prevent overfitting, we set a relatively low learning rate of $8 \times 10^{-7}$ and adopted a cosine learning rate scheduler for learning rate decay. During training, the per-device batch size was set to 4 with gradient accumulation steps of 1, training for a total of 1 epoch. We utilized DeepSpeed Zero-3 optimization strategy to handle the memory requirements of large models.

\textbf{Reinforcement Learning (RL) Training Stage:} Following SFT completion, we employed the EasyR1(using VERL) framework for reinforcement learning training. The training dataset contained 15k samples, with the geometry3k test set used for validation. Key RL training configurations included: rollout batch size set to 128, with 12 rollouts generated per sample for policy optimization. The Actor model's global batch size was 128, with micro batch size per device for updates set to 2 and micro batch size per device for experience collection set to 4.  The optimizer used was AdamW, and the learning rate was set to $1 \times 10^{-6}$. To maintain visual feature stability, we froze the vision tower parameters and fine-tuned only the language components.

\section{Prompt Template}
\label{app:Prompt_Template}
We provide the detailed prompt templates used for generating and training the \texttt{<back>} insertion and reasoning behaviors within Look-Back. These prompts were designed to guide the model to autonomously revisit and verify visual information during the reasoning process in a structured manner, without explicit re-injection of images.

Specifically, we design separate prompt templates for:
\begin{itemize}[nolistsep, leftmargin=*]
\item \textbf{Semantic-back:} inserting \texttt{<back>} within the reasoning chain to verify intermediate reasoning steps against the image while allowing the model to continue its reasoning.
\item \textbf{Solution-back:} inserting \texttt{<back>} after completing an initial reasoning chain to trigger a comprehensive rethinking process based on image verification.
\end{itemize}

We provide templates tailored for both Supervised Fine-Tuning (SFT) and Reinforcement Learning (RL) settings, ensuring consistent output structures and stable triggering of the \texttt{<back>} mechanism. These templates enforce a structured output format:

\begin{promptbox*}[Prompt Template for Semantic-back Insertion]{lightblue}{prompt:back-insertion}
\label{temp:semantic-back-insertion-template}

\textbf{User:} 
\begin{quote}
You are an expert at completing Chain-of-Thought (CoT) reasoning processes. You will be given:

\begin{itemize}
    \item An image
    \item A user prompt (the question)
    \item A CoT reasoning process to the current question from another model (\textbf{not} contain any \texttt{<back>} tags)
    \item The correct answer to the question
\end{itemize}

Your task is to insert \texttt{<back> ... </back>} tags at appropriate locations in the reasoning process based on the correctness of the model's answer, simulate a ``review" of the image, and verify whether the reasoning is consistent with the visual content.

\textbf{Rules for inserting \texttt{<back>}:}

\begin{enumerate}
    \item If the model's answer is wrong, you should check where the error is in the reasoning, and point out the logical errors, missing picture assumptions or information (combined with the picture) in the \texttt{<back>}, and modify the subsequent reasoning process and answer according to the content. The reasoning before the error step cannot be modified.
    
    \item If the model's answer is correct, you can insert \texttt{<back>} appropriately to verify the reasoning related to the picture. The reasoning process is not modified, just insert \texttt{<back>}.
    
    \item The content in \texttt{<back>} must be \textbf{succinct and clear}, preferably \textbf{one sentence (excluding filler phrases)}, used to verify the reasoning related to the picture.
    
    \item The final format should look like this:
    \begin{verbatim}
<think>
Sentence 1 of reasoning.
Sentence 2 of reasoning. </think>
<back> Image-based verification goes here. </back>
<think>
Further reasoning. </think>
\boxed{final answer}
    \end{verbatim}
\end{enumerate}

\end{quote}

\texttt{--- Input ---} \\
\texttt{Question:} \{user\_prompt\} \\
\texttt{CoT Response:} \{prediction\} \\
\texttt{Correct Answer:} \{answer\} \\
\texttt{Score:} \{score\}

\texttt{--- Output Format ---} \\
The output format is: \texttt{<think>The reasoning process is here </think> <back>The verification process is here </back> <think>Continue reasoning </think> \textbackslash boxed\{answer\}}
\end{promptbox*}

\begin{promptbox*}[Prompt Template for Solution-back Insertion]{lightblue}{prompt:back-rethinking}
\label{temp:solution-back-rethinking-template}

\textbf{User:} 
\begin{quote}
You are an expert at completing Chain-of-Thought (CoT) reasoning processes. You will be given:

\begin{itemize}
    \item An image
    \item A user prompt (the question)
    \item A CoT reasoning process to the current question from another model (\textbf{not} contain any \texttt{<back>} tags)
    \item The correct answer to the question
\end{itemize}

Your task is to insert \texttt{<back> ... </back>} tags after \texttt{</think>} according to the correctness of the model's answer, simulate a ``review" of the image, and verify whether the reasoning is consistent with the visual content.

\textbf{Rules for inserting \texttt{<back>}:}

\begin{enumerate}
    \item If the model's answer is wrong, it should examine where the error is in the reasoning and point out the logical error, missing picture assumptions, or information (with pictures) in \texttt{<back>}. And rethink and revise the reasoning and answer based on the content after \texttt{</back>}.
    
    \item If the model's answer is correct, only insert \texttt{<back>} to verify the reasoning related to the image. And rethink and give the answer after \texttt{</back>}.
    
    \item The content in \texttt{<back>} must be \textbf{succinct and clear}, used to verify the reasoning related to the picture.

    \item The final format should look like this:
    \begin{verbatim}
<think>
Sentence 1 of reasoning.
Sentence 2 of reasoning. </think>
<back> Image-based verification goes here. </back>
<think>
based on the thinking and verification contents, rethinking </think>
\boxed{final answer}
    \end{verbatim}
\end{enumerate}

\textbf{Important Notes:}
\begin{itemize}
    \item Do not modify the content of \texttt{<think>} in CoT itself when inserting \texttt{<back>}, just add it later.
    \item If you find any errors in your reasoning or inconsistencies with the picture, please correct them in \texttt{<back>} and modify the subsequent reasoning and answers.
\end{itemize}
\end{quote}

\texttt{--- Input ---} \\
\texttt{Question:} \{user\_prompt\} \\
\texttt{CoT Response:} \{prediction\} \\
\texttt{Correct Answer:} \{answer\} \\
\texttt{Score:} \{score\}

\texttt{--- Output Format ---} \\
The output format is: \texttt{<think> reasoning process here </think> <back> verification process against the image here </back> <think> based on the thinking and verification contents, rethinking here </think> \textbackslash boxed\{answer\}}
\end{promptbox*}

\begin{promptbox}[Prompt Template for Semantic-back RL]{lightblue}{prompt:Semantic-back-RL}
\label{temp:Semantic-back RL}

\textbf{System:}
You are a helpful assistant.

\vspace{0.5em}
\textbf{User:}
\{Query\} You FIRST think about the reasoning process as an internal monologue and then provide the final answer. The reasoning process MUST BE enclosed within \texttt{<think> </think>} tags, and use \texttt{<back> </back>} to verify your reasoning against the image. The final answer MUST BE put in \texttt{\textbackslash boxed\{\}}, respectively, i.e., \texttt{<think> reasoning process here </think> <back> verification process here </back> <think> continue reasoning </think> \textbackslash boxed\{final answer\}}.
\end{promptbox}

\begin{promptbox}[Prompt Template for Solution-back RL]{lightblue}{prompt:Solution-back-RL}
\label{temp:Solution-back RL}

\textbf{System:}
You are a helpful assistant.

\vspace{0.5em}
\textbf{User:}
\{Query\} You FIRST think about the reasoning process as an internal monologue and then provide the final answer. The reasoning process MUST BE enclosed within \texttt{<think> </think>} tags, and use \texttt{<back> </back>} to verify your reasoning solution against the image, and rethink it in the \texttt{<think> </think>} tags based on your thinking and verification content. The final answer MUST BE put in \texttt{\textbackslash boxed\{\}}, respectively, i.e., \texttt{<think> reasoning process here </think> <back> verification process against the image here </back> <think> based on the thinking and verification contents, rethinking here </think> \textbackslash boxed\{final answer\}}.
\end{promptbox}

\begin{promptbox}[Prompt Template for Solution-back SFT]{lightblue}{prompt:Solution-back-SFT}
\label{temp:Solution-back SFT}

\textbf{System:} You are a helpful assistant.
\vspace{0.5em}

\textbf{User:}
\begin{verbatim}
{
  "messages": [
    {
      "role": "user",
      "content": "<image> {query} You FIRST think
      about the reasoning process as an internal 
      monologue and then provide the final answer.
      The reasoning process MUST BE enclosed within 
      <think> </think> tags, and use <back> </back> 
      to verify your reasoning solution against the 
      image, and rethink it in the <think> </think> 
      tags based on your thinking and verification
      content. The final answer MUST BE put in
      \\boxed{}, respectively, i.e., <think> 
      reasoning process here </think> <back> 
      verification process against the image here 
      </back> <think> based on the thinking and 
      verification contents, rethinking here 
      </think> \\boxed{final answer}."
    },
    {
      "role": "assistant",
      "content": "CoT with <back>"
    }
  ],
  "images": ["{image_url}"]
}
\end{verbatim}

\end{promptbox}

\begin{promptbox}[Prompt Template for Semantic-back SFT]{lightblue}{prompt:Semantic-back-SFT}
\label{temp:Semantic-back SFT}

\textbf{System:} You are a helpful assistant.
\vspace{0.5em}

\textbf{User:}
\begin{verbatim}
{
  "messages": [
    {
      "role": "user",
      "content": "<image> {query} You FIRST think 
      about the reasoning process as an internal  
      monologue and thenprovide the final answer.  
      The reasoning process MUST BE enclosed 
      within <think> </think> tags, and use <back> 
      </back> to verify your reasoning against the 
      image.The final answer MUST BE put in \\boxed{}, 
      respectively, i.e., <think> reasoning process
      here </think> <back> verification process here
      </back> <think> continue reasoning </think> 
      \\boxed{final answer}."
    },
    {
      "role": "assistant",
      "content": "CoT with <back>"
    }
  ],
  "images": ["{image_url}"]
}
\end{verbatim}

\end{promptbox}

\section{Case Study}
\label{app: Case_study}
\textbf{Selected Benchmark Cases.}  
We curate five representative instances drawn from \emph{MathVision}, \emph{MME}, \emph{HalluBench}, \emph{MathVerse}, and \emph{TallyQA}.  
Samples~1--3 are generated by the \emph{Solution-back} variant, which first completes an entire Chain-of-Thought (CoT) and then invokes a final \texttt{<back>} segment to re-inspect the image.  
Samples~4--5 originate from the \emph{Semantic-back} model, where \texttt{<back>} is triggered \emph{during} reasoning, allowing the model to intermittently verify visual details before continuing the CoT.  
These two backtracking modes therefore span both post-solution review (Solution-back) and in-process visual reflection (Semantic-back) across diverse mathematical and perceptual benchmarks.

\textbf{Effect of Visual Reflection.}  
Despite differing trigger timings, all five examples exhibit the same corrective pattern: the initial \texttt{<think>} segment contains a flawed deduction, which is subsequently examined against the image inside \texttt{<back>}.  
By explicitly grounding its reasoning in visual evidence—whether counting shaded areas, identifying a landmark church, discerning rotation direction, resolving arc measures, or spotting a single bird—the model revises the error and outputs the ground-truth answer.  
This qualitative study highlights how both \emph{Solution-back} and \emph{Semantic-back} reliably leverage self-reflection on visual cues to transform incorrect intermediate reasoning into accurate final predictions.

\begin{promptbox*}[Generated Samples of Solution-back in MathVision (Sample 1)]{lightorange}{prompt:Sample_1}
\noindent
\begin{minipage}[c]{0.37\textwidth}
  \centering
  \includegraphics[width=\linewidth]{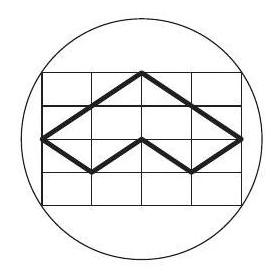}
\end{minipage}
\hfill
\begin{minipage}[c]{0.60\textwidth}
  \textbf{Question}: 
  The diagram shows five circles of the same radius touching each other.  
  A square is drawn so that its vertices are at the centres of the four outer circles.  
  What is the ratio of the area of the shaded parts of the circles to the area of the unshaded parts of the circles?\\[4pt]

\noindent
\colorbox{gray!15}{%
  \parbox{\linewidth}{\raggedright\ttfamily
    Source: MathVision, id:1723\\
    Model: Solution-back%
  }%
}
\\

  \textbf{Choices}: (A) $1:3$ \quad (B) $1:4$ \quad (C) $2:5$ \quad (D) $2:3$ \quad (E) $5:4$\\
  \textbf{\textcolor{red}{Ground truth: (D) $2:3$}}
\end{minipage}

\vspace{6pt}

\begingroup\ttfamily
\textbf{<think>} \\
To solve this problem, we need to determine the ratio of the area of the shaded parts of the circles to the area of the unshaded parts of the circles.

1. Identify the Geometry:\\
\quad-- There are five circles of the same radius \( r \).\\
\quad-- A square is drawn such that its vertices are at the centres of the four outer circles.\\
\quad-- The shaded parts are the areas of the circles that are not covered by the square.\\
2. Calculate the Area of the Square:\\
\quad-- The side length of the square is equal to the diameter of two circles, which is \( 2r \).\\
\quad-- The area of the square is \( (2r)^2 = 4r^2 \).\\
3. Calculate the Area of the Circles:\\
\quad-- The area of one circle is \( \pi r^2 \).\\
\quad-- Since there are five circles, the total area of the circles is \( 5\pi r^2 \).\\
4. Determine the Shaded and Unshaded Areas:\\
\quad-- The shaded parts are the areas of the circles that are not covered by the square.\\
\quad-- The unshaded parts are the areas of the circles that are covered by the square.\\
5. Calculate the Ratio:\\
\quad-- The area of the shaded parts is the total area of the circles minus the area of the square.\\
\quad-- The area of the unshaded parts is the area of the square.\\
6. Simplify the Ratio:\\
\quad-- The ratio of the area of the shaded parts to the area of the unshaded parts is\\
\quad\quad\( \dfrac{5\pi r^2 - 4r^2}{4r^2} \).\\

After simplifying, we find that the ratio is \( 1:3 \). \textcolor{red}{(Incorrect)}\\
\textbf{</think>} \\[4pt]
\textbf{<back>} \\
Verification against the image shows that the four outer circles each contribute a quarter-circle area \( \tfrac14\pi r^2 \) to the shaded region, and the centre circle contributes a full-circle area \( \pi r^2 \).\\
Hence shaded area \( = 4 \times \tfrac14\pi r^2 + \pi r^2 = 2\pi r^2 \).\\
Unshaded area \( = 5\pi r^2 - 2\pi r^2 = 3\pi r^2 \).\\
Therefore the correct shaded : unshaded ratio is \( 2:3 \), not \( 1:3 \).\\
\textcolor{green}{(Corrected using the image; correct answer obtained)}\\
\textbf{</back>} \\[4pt]
\textbf{<think>} \\
Re-examining the geometry with the corrected areas confirms the ratio is indeed \( 2:3 \).\\
\textbf{</think>}
\(\boxed{D}\) \textcolor{green}{Correct}
\endgroup

\end{promptbox*}

\begin{promptbox*}[Generated Samples of Solution-back in MME (Sample 2)]{lightorange}{prompt:Sample_2}
\noindent
\begin{minipage}[c]{0.37\textwidth}
  \centering
  \includegraphics[width=\linewidth]{fig/MME.jpg}
\end{minipage}
\hfill
\begin{minipage}[c]{0.60\textwidth}
  \textbf{Question}: 
  Is this an image of \emph{Church of Saint John the Baptist (Jihlava)}?  Please answer \textbf{yes} or \textbf{no}.\\[4pt]

\noindent
\colorbox{gray!15}{%
  \parbox{\linewidth}{\raggedright\ttfamily
    Source: Landmark, id:landmark/1ac0e3c493073995\\
    Model: Solution-back%
  }%
}
\\

  \textbf{\textcolor{red}{Ground truth: Yes}}
\end{minipage}

\vspace{6pt}

\begingroup\ttfamily
\textbf{<think>} \\
The image shows a small, white church with a dark roof and a bell tower. The architecture and the surrounding greenery are consistent with a small, rural church. The Church of Saint John the Baptist in Jihlava is known for its specific architectural features, which include a more elaborate design with a larger bell tower and a different roof style. The church in the image does not match the known architectural features of the Church of Saint John the Baptist in Jihlava. \textcolor{red}{(Incorrect)}\\
\textbf{</think>} \\[4pt]
\textbf{<back>} \\
Verification against the image shows a small, simple church with a single bell tower and a dark, sloped roof, which is consistent with the description of the Church of Saint John the Baptist in Jihlava. The image does not show the more elaborate features of a larger church, such as additional towers or a more complex roof structure. \textcolor{green}{(Corrected using the image; correct answer obtained)}\\
\textbf{</back>} \\[4pt]
\textbf{<think>} \\
Based on this verification, the church in the image is consistent with the known features of the Church of Saint John the Baptist in Jihlava.\\
\textbf{</think>}
\(\boxed{Yes}\) \textcolor{green}{Correct}
\endgroup

\end{promptbox*}

\begin{promptbox*}[Generated Samples of Solution-back in Hallubench (Sample 3)]{lightorange}{prompt:Sample_3}
\noindent
\begin{minipage}[c]{0.37\textwidth}
  \centering
  \includegraphics[width=\linewidth]{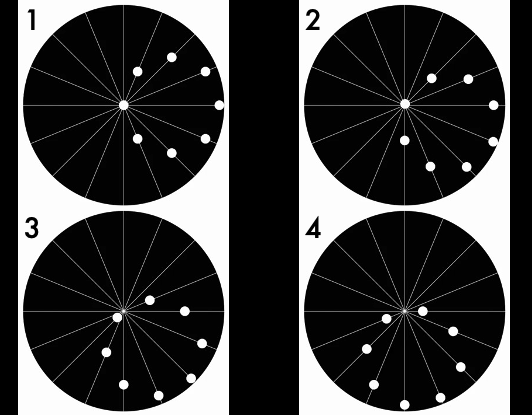}
\end{minipage}
\hfill
\begin{minipage}[c]{0.60\textwidth}
  \textbf{Question}: 
  According to the positive sequence of the images, is the circle inside autobiographical anticlockwise?  
  Please answer \textbf{yes} or \textbf{no}.\\[4pt]

\noindent
\colorbox{gray!15}{%
  \parbox{\linewidth}{\raggedright\ttfamily
    Source: Hallubench, set\_id: 9, figure\_id: 1, question\_id: 0\\
    Model: Solution-back%
  }%
}
\\

  \textbf{\textcolor{red}{Ground truth: no}}
\end{minipage}

\vspace{6pt}

\begingroup\ttfamily
\textbf{<think>} \\
To determine if the circle inside the autobiographical anticlockwise, we need to analyze the sequence of images and the direction of the dots.

1. The first image shows a circle with dots placed in a specific pattern.\\
2. The second image shows the same circle with the dots shifted in a certain direction.\\
3. The third image shows the dots further shifted in the same direction.\\
4. The fourth image shows the dots shifted again in the same direction.\\

By observing the sequence, we can see that the dots are moving in a consistent direction. If we assume the direction is anticlockwise, we need to check if the final position of the dots matches the expected anticlockwise pattern.
Given the sequence, the dots appear to be moving in an anticlockwise direction. Therefore, the circle inside is indeed moving in an anticlockwise direction. \textcolor{red}{(Incorrect)}\\
\textbf{</think>} \\[4pt]
\textbf{<back>} \\
Verification against the image shows that the dots are consistently moving from the top clockwise to the bottom left, then to the bottom right, and finally to the top right, which is the opposite of anticlockwise. The dots are actually moving in a clockwise direction. \textcolor{green}{(Corrected using the image; correct answer obtained)}\\
\textbf{</back>} \\[4pt]
\textbf{<think>} \\
Re-examining the sequence, the dots are clearly moving in a clockwise direction, not anticlockwise.\\
\textbf{</think>}
\(\boxed{no}\) \textcolor{green}{Correct}
\endgroup

\end{promptbox*}

\begin{promptbox*}[Generated Samples of Semantic-back in MathVerse (Sample 4)]{lightorange}{prompt:Sample_4}
\noindent
\begin{minipage}[c]{0.37\textwidth}
  \centering
  \includegraphics[width=\linewidth]{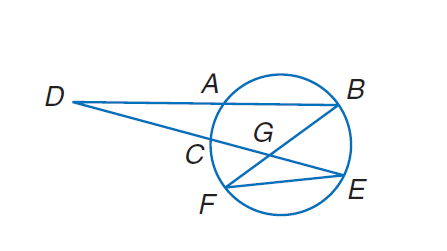}
\end{minipage}
\hfill
\begin{minipage}[c]{0.60\textwidth}
  \textbf{Question}: 
  \textbf{D} is an external point to the circle, and \textbf{A, C, F, E, B} are on the circle. If $m\widehat{FE}=118$, $m\widehat{AB}=108$, $m\angle EGB=52$, and $m\angle EFB=30$, find $m\widehat{CF}$.\\[4pt]

\noindent
\colorbox{gray!15}{%
  \parbox{\linewidth}{\raggedright\ttfamily
    Source: MathVerse, id:1231\\
    Model: Semantic-back%
  }%
}
\\

  \textbf{Choices}: (A) 30 \quad (B) 44 \quad (C) 108 \quad (D) 118\\
  \textbf{\textcolor{red}{Ground truth: (B) 44}}
\end{minipage}

\vspace{6pt}

\begingroup\ttfamily
\textbf{<think>} \\
To find the measure of the arc \(\widehat{CF}\), we need to use the given information and properties of circles and angles. Let's break it down step by step.

1. Identify the given information:\\
\quad-- \(m \widehat{FE} = 118^\circ\)\\
\quad-- \(m \widehat{AB} = 108^\circ\)\\
\quad-- \(m \angle EGB = 52^\circ\)\\
\quad-- \(m \angle EFB = 30^\circ\)\\
2. Understand the relationship between the angles and arcs:\\
\quad-- The measure of an inscribed angle is half the measure of the arc it intercepts.\\
\quad-- The measure of an angle formed by a tangent and a chord is equal to half the measure of the intercepted arc.\\
3. Determine the measure of \(\angle EGB\):\\
\quad-- \(\angle EGB\) is an inscribed angle that intercepts arc \(\widehat{EB}\).\\
\quad-- Therefore, \(m \widehat{EB} = 2 \times m \angle EGB = 2 \times 52^\circ = 104^\circ\).\\
4. Determine the measure of \(\angle EFB\):\\
\quad-- \(\angle EFB\) is an inscribed angle that intercepts arc \(\widehat{EB}\).\\
\quad-- Therefore, \(m \widehat{EB} = 2 \times m \angle EFB = 2 \times 30^\circ = 60^\circ\).\\
5. Find the measure of \(\widehat{CF}\):\\
\quad-- The measure of \(\widehat{CF}\) is the difference between the measure of \(\widehat{FE}\) and the measure of \(\widehat{EB}\).\\
\quad-- Therefore, \(m \widehat{CF} = m \widehat{FE} - m \widehat{EB} = 118^\circ - 60^\circ = 58^\circ\).\\
\textcolor{red}{(Incorrect)}
\textbf{</think>} \\[4pt]
\textbf{<back>} \\
From the diagram, \(\angle EGB\) intercepts arc \(\widehat{EB}\), so \(m \widehat{EB} = 2 \times 52^\circ = 104^\circ\), not \(60^\circ\). \textcolor{green}{(Corrected using the image; correct answer obtained)}
\textbf{</back>} \\[4pt]
\textbf{<think>} \\
6. Correct the measure of \(\widehat{EB}\):\\
\quad-- The correct measure of \(\widehat{EB}\) is \(104^\circ\), not \(60^\circ\).\\
7. Recompute the measure of \(\widehat{CF}\):\\
\quad-- Therefore, \(m \widehat{CF} = m \widehat{FE} - m \widehat{EB} = 118^\circ - 104^\circ = 14^\circ\).\\
\textbf{</think>} 
\(\boxed{B}\) \textcolor{green}{Correct}
\endgroup

\end{promptbox*}

\begin{promptbox*}[Generated Samples of Semantic-back in Tallyqa (Sample 5)]{lightorange}{prompt:Sample_5}
\noindent
\begin{minipage}[c]{0.37\textwidth}
  \centering
  \includegraphics[width=\linewidth]{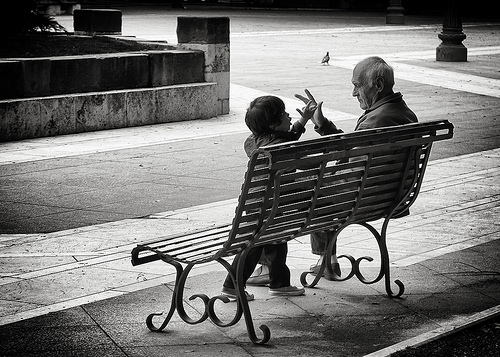}
\end{minipage}
\hfill
\begin{minipage}[c]{0.60\textwidth}
  \textbf{Question}: 
  How many birds are on the sidewalk?\\[4pt]

\noindent
\colorbox{gray!15}{%
  \parbox{\linewidth}{\raggedright\ttfamily
    Source: Tallyqa, id:945\\
    Model: Semantic-back%
  }%
}
\\

  \textbf{\textcolor{red}{Ground truth: 1}}
\end{minipage}

\vspace{6pt}

\begingroup\ttfamily
\textbf{<think>} \\
To answer the question, I need to carefully observe the image for any birds on the sidewalk. I'll scan the entire scene, including the bench, the people, and the background. There are no birds visible on the sidewalk. The image shows an elderly man and a child sitting on a bench, and the background includes some steps and a small area that appears to be a park or plaza. There are no birds in the sky or on the ground. \textcolor{red}{(Incorrect)}\\
\textbf{</think>} \\[4pt]
\textbf{<back>} \\
There is a small bird visible on the ground in the background, near the steps. \textcolor{green}{(Corrected using the image; correct answer obtained)}\\
\textbf{</back>} \\[4pt]
\textbf{<think>} \\
After re-evaluating, I see a small bird on the ground in the background, near the steps.\\
\textbf{</think>} \(\boxed{1}\)  \textcolor{green}{Correct}
\endgroup
\end{promptbox*}

\section{Training Dynamics}
\label{app: Training_Dynamics}
\begin{figure*}[htbp]
  \centering
  \begin{subfigure}[b]{0.48\linewidth}
    \includegraphics[width=\linewidth]{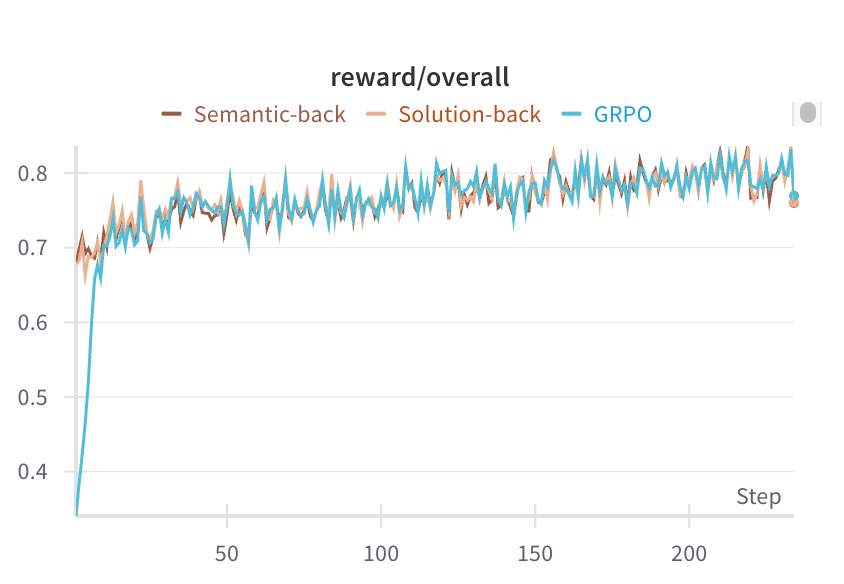}
  \end{subfigure}
  \hfill
  \begin{subfigure}[b]{0.48\linewidth}
    \includegraphics[width=\linewidth]{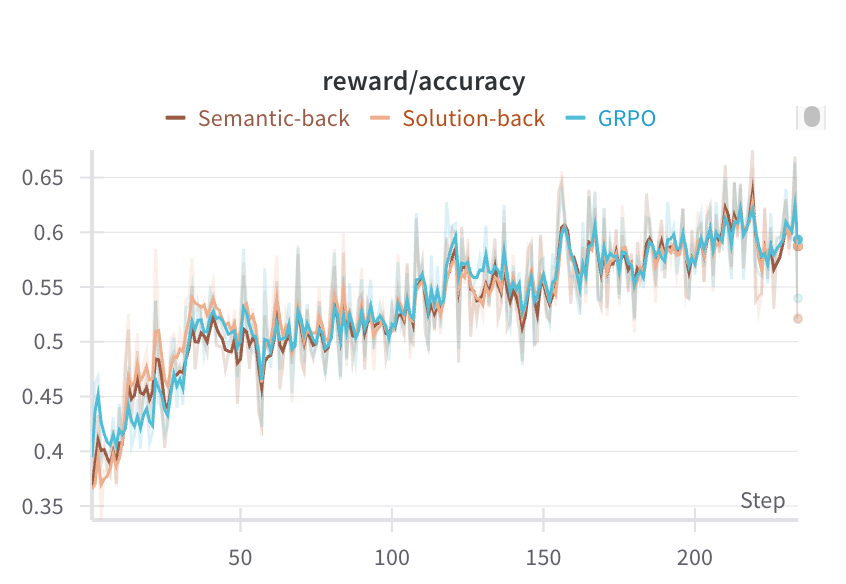}
  \end{subfigure}

  \vspace{0.5em}

  \begin{subfigure}[b]{0.48\linewidth}
    \includegraphics[width=\linewidth]{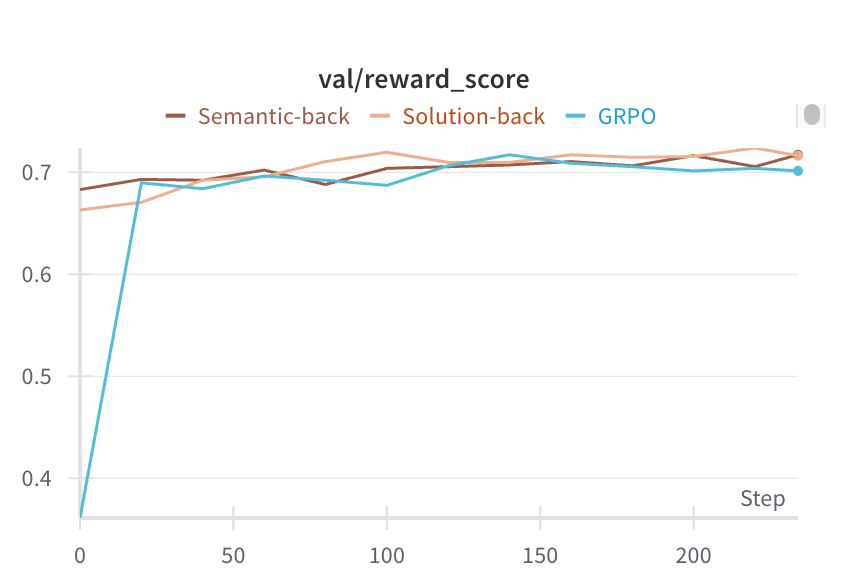}
  \end{subfigure}
  \hfill
  \begin{subfigure}[b]{0.48\linewidth}
    \includegraphics[width=\linewidth]{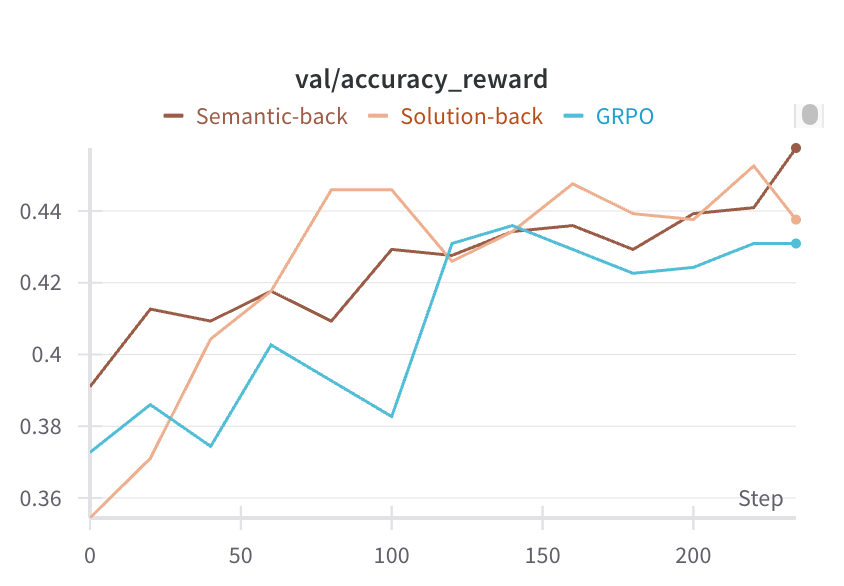}
  \end{subfigure}

  \vspace{0.5em}

  \begin{subfigure}[b]{0.48\linewidth}
    \includegraphics[width=\linewidth]{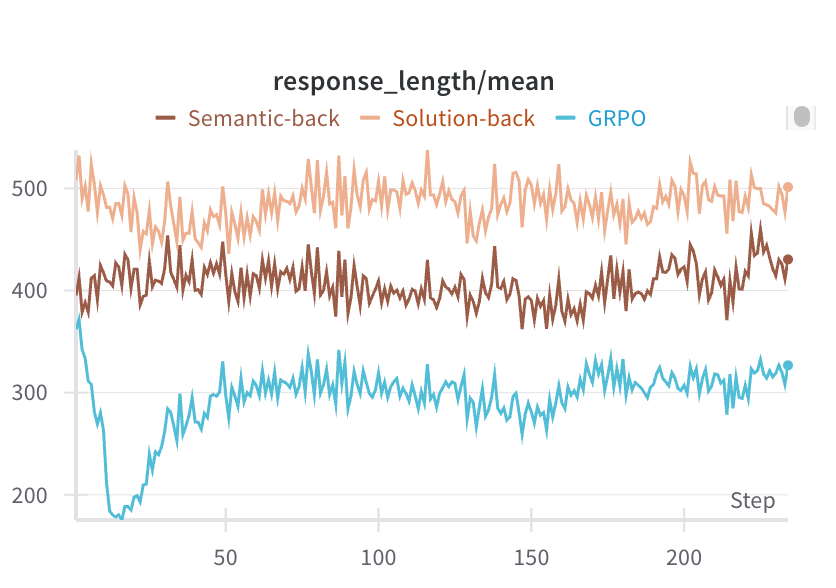}
  \end{subfigure}
  \hfill
  \begin{subfigure}[b]{0.48\linewidth}
    \includegraphics[width=\linewidth]{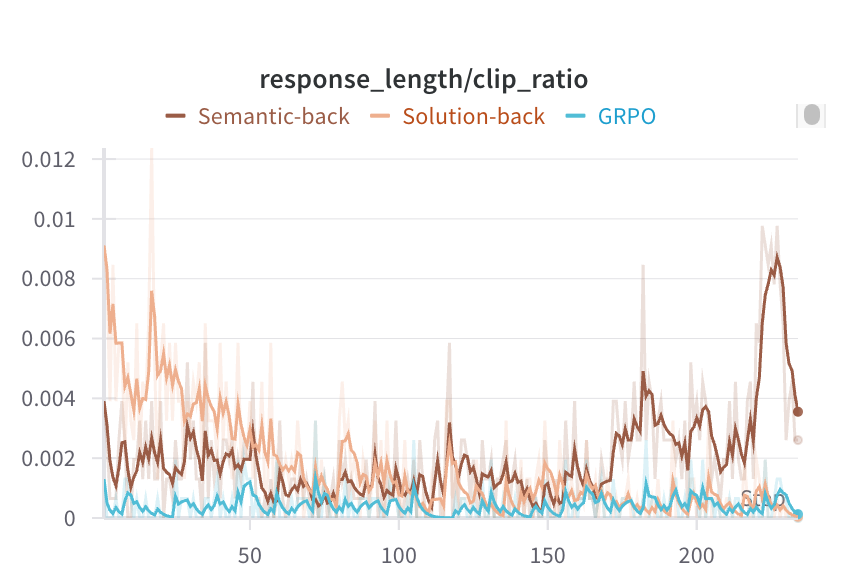}
  \end{subfigure}

  \caption{More training dynamics of GRPO, Semantic-back and Solution-back.}
  \label{fig:six_plots}
\end{figure*}

We present additional training dynamics for GRPO, Semantic-back, and Solution-back, as shown in Figure \ref{fig:six_plots}. The visualizations include reward (on both training and validation sets), accuracy, response length, and clip ratio across training steps.

\end{document}